%% file: main.tex
\documentclass[runningheads]{llncs}

\usepackage[T1]{fontenc}

\usepackage{mathrsfs}
\newcommand{\argmin}{\mathop{\mathrm{arg\,min}}\limits}
\usepackage{amsmath, amssymb}
\usepackage{booktabs}
\usepackage{longtable}
\usepackage{graphicx,subcaption}
\usepackage{tikz}
\usetikzlibrary{calc}
\usepackage{array}
\usepackage{wrapfig}
\usepackage{marvosym}
\usepackage{makecell}
\usepackage{color}
\usepackage{hyperref}
\usepackage{tcolorbox}
\tcbuselibrary{skins}

\newcommand{\equalcontrib}{\textsuperscript{*}}
\newcommand{\corrauth}{\textsuperscript{\Letter}}
\newcommand{\roundedimg}[2][4pt]{%
  \begin{tikzpicture}
    \clip[rounded corners=#1] (0,0) rectangle (\linewidth,\linewidth);
    \node[anchor=south west, inner sep=0] at (0,0)
      {\includegraphics[width=\linewidth]{#2}};
  \end{tikzpicture}%
}

\begin{document}

\title{MULTI: Disentangling Camera Lens, Sensor, View, and Domain for Novel Image Generation}

\titlerunning{\textbf{MULTI}: Disentangling Imaging Factors}

\author{%
  Sonali Godavarthy\inst{1,3}\equalcontrib
  \and
  Matthias Neuwirth-Trapp\inst{1,2}\equalcontrib\corrauth
  \and
  Tim-Felix Faasch\inst{1}
  \and
  Maarten Bieshaar\inst{1}
  \and
  Michael Moeller\inst{3}
  \and
  Danda Pani Paudel\inst{4}
}

\authorrunning{S. Godavarthy, M. Neuwirth-Trapp et al.}

\institute{Bosch Research, Hildesheim, Germany \and
ETH Zürich, Zürich, Switzerland \and
University of Siegen, Siegen, Germany \and
INSAIT, Sofia University “St.~Kliment Ohridski”, Sofia, Bulgaria}

\maketitle

\begingroup
\renewcommand\thefootnote{}
\footnotetext{\textsuperscript{*}These authors contributed equally to this work.\\
\textsuperscript{\Letter}Corresponding author: Matthias Neuwirth-Trapp, 
\href{mailto:mneuwirth@ethz.ch}{mneuwirth@ethz.ch}.}
\endgroup

\input{sec/0_abstract}

\input{sec/1_intro}
\input{sec/2_related_works}
\input{sec/3_problem}
\input{sec/4_method}
\input{sec/5_experiments}

\input{sec/6_analysis}
\input{sec/7_conclusion}

\vspace{-0.3cm}
\subsubsection{Acknowledgments} 
Sonali Godavarthy and  Michael Moeller acknowledge the support of the DFG Research Unit ``Learning to Sense'', project 459284860.

{
    \bibliographystyle{splncs04}
    \bibliography{main, first-paper}
}

\input{supplementary}

\end{document}

%% file: sec/0_abstract.tex
\vspace{-0.5cm}
\begin{abstract} Recent text-to-image models produce high-quality images, yet text ambiguity hinders precise control when specific styles or objects are required. There have been a number of recent works dealing with learning and composing multiple objects and patterns. However, current work focuses almost entirely on image content, overlooking imaging factors such as camera lens, sensor types, imaging viewpoints, and scenes' domain characteristics. We introduce this new challenge as \emph{Imaging Factor Disentanglement} and show limitations of current approaches in the regime. We, therefore, propose the new method \textbf{Mul}ti-factor disentanglement through \textbf{T}extual \textbf{I}nversion (\textbf{MULTI}). It consists of two stages: in the first stage, we learn general factors, and in the second stage, we extract dataset-specific ones. This setup enables the extension of existing datasets and novel factor combinations, thereby reducing distribution gaps. It further supports modifications of specific factors and image-to-image generation via ControlNets. The evaluation on our new DF-RICO benchmark demonstrates the effectiveness of MULTI and highlights the importance of Factor Disentanglement as a new direction of research.

\keywords{Text-to-Image \and Diffusion Models \and Factor Disentanglement~\and Textual Inversion \and Controllable Generation \and Stable Diffusion}

\end{abstract}
\vspace{-0.5cm}

%% file: sec/1_intro.tex
\begin{figure}[t]
    \centering
    \includegraphics[width=1\linewidth]{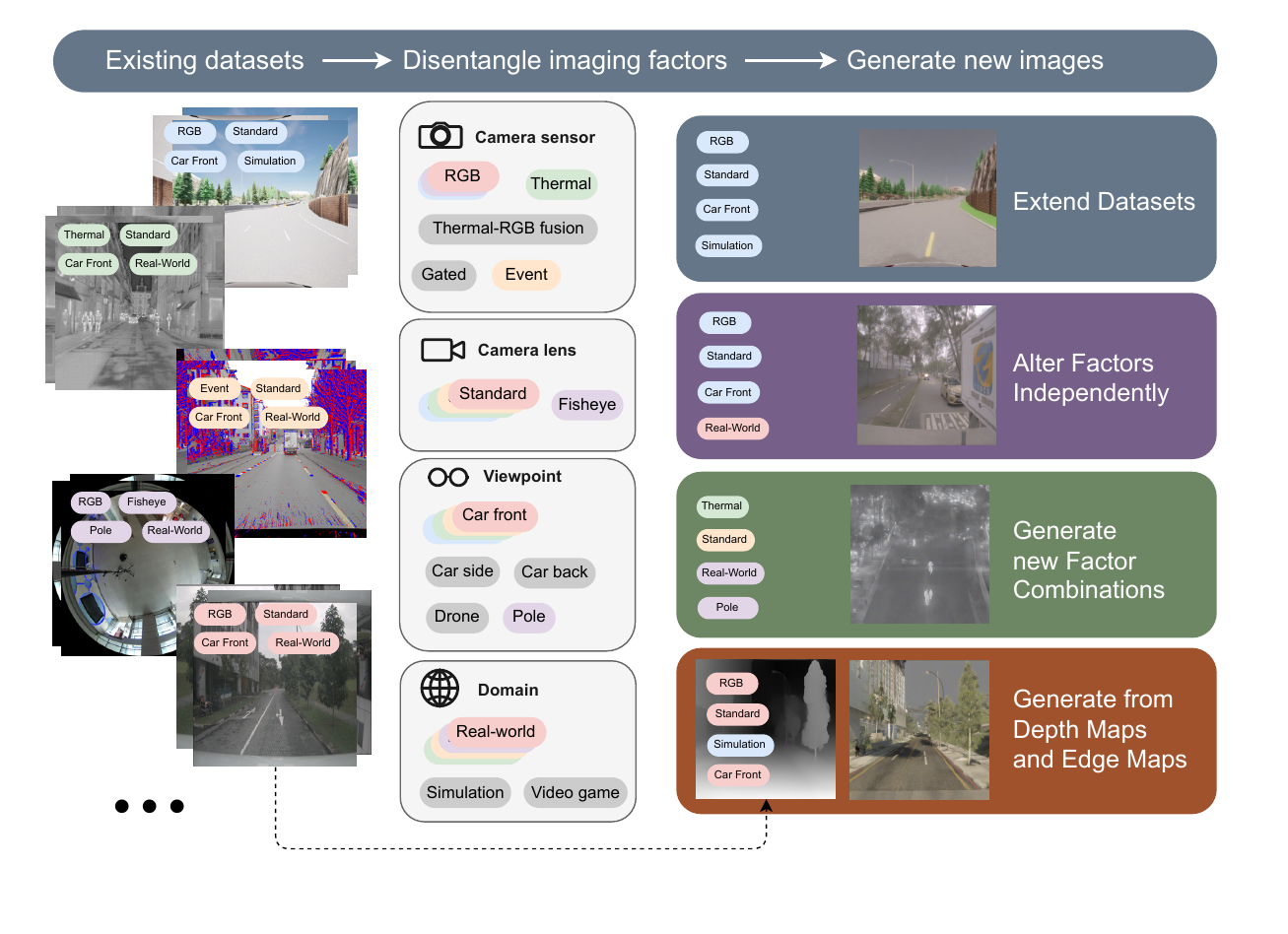}
    \caption{\textbf{Overview of Factor Disentanglement.} In this work, we propose the challenge of \emph{Imaging Factor Disentanglement}, namely, the camera lens and sensor types with associated color grading, viewpoint, and domain from a set of sparse and unpaired datasets. }
    \label{fig:fig1}
\end{figure}

\section{Introduction}
\label{sec:introduction}

State-of-the-art (SOTA) Text-to-Image (T2I) models have recently achieved a level of visual fidelity that nearly matches real-world imagery~\cite{esser2024scaling,nichol2022glide,podell2023sdxl,rombach2022high,saharia2022photorealistic}. Despite this progress, relying entirely on text prompts remains insufficient for many practical applications~\cite{gal2022image,ruiz2023dreambooth,zhang2023adding}. Natural language is inherently limited in describing the nuances of abstract concepts. The factors that govern the physical image formation process, such as specific lens distortions, sensor noise profiles, or exact viewpoints, cannot be reliably encoded through text alone. 

The visual appearance of an image is mainly governed by four key factors: the camera lens, the sensor and its associated color profile, the viewpoint, and the data domain. Precise control over these factors is critical for the application of diffusion models as task-specific simulations. Although approaches like LoRA~\cite{hu2022lora}, DreamBooth~\cite{ruiz2023dreambooth}, and ControlNets~\cite{zhang2023adding} have improved controllability, they primarily focus on high-level concepts such as object identity or artistic style~\cite{butt2024colorpeel,vinker2023concept,xu2024dreamanime}. They do not address the low-level imaging characteristics.

In this context, we formally address the challenge of \emph{Imaging Factor Disentanglement} for the first time. Our goal is to control imaging characteristics independently of the scene content. We aim to control both the general type of \textbf{factor} (e.g., differentiating a fisheye lens from a standard rectilinear lens) and its \textbf{specific factor} (e.g., a specific lens distortion pattern). Solving this challenge allows for the extension of datasets beyond their original capture conditions by enabling the synthesis of novel factor combinations that fill the missing regions of the imaging factor distribution.

To address the challenge, we introduce \textbf{MULTI} (\textbf{Mul}ti-factor disentanglement through \textbf{T}extual \textbf{I}nversion). MULTI is the first approach designed to obtain control over imaging factors using pre-trained T2I models. We encode the factors into learnable embedding vectors that are combined with descriptive prompts to guide the image generation. We handle the sparsity of real-world data by employing a novel two-stage training approach. In the first stage, we leverage partially overlapping datasets to learn general embeddings. We select training batches in which the samples share common factors but come from different datasets and isolate common imaging characteristics through optimization of the embeddings. In the second stage, we induce dataset-specific knowledge, allowing the model to replicate exact sensor or lens profiles. This setup is efficient to train and fully compatible with ControlNets, allowing for precise spatial control alongside imaging factor control, at the same time. This allows the extension of the datasets, the generation of images with novel combinations, and enables Image-to-Image (I2I) generation with ControlNets as seen in Fig.~\ref{fig:fig1}.

To rigorously study this new challenge, we adapt the RICO benchmark~\cite{neuwirth2025rico}, which spans varied imaging factor configurations, to \textbf{DF-RICO} (\textbf{D}isentangling \textbf{F}actors with \textbf{RICO}). We show that our approach successfully disentangles the imaging factors, allowing the generation of new images with existing or novel imaging factor configurations while optionally preserving the spatial layout of an existing image. Our main contributions are:
\begin{itemize} 
\item \textbf{Imaging Factor Disentanglement:} We formalize the challenge of disentangling low-level imaging factors, e.g., lens, sensor, viewpoint, and domain, into physically interpretable primitives from sparse data, addressing the limitations of natural language in describing image formation. 
\item \textbf{The MULTI Approach:} We introduce MULTI, a novel two-stage textual inversion strategy that exploits partially overlapping datasets to isolate and recombine independent factors without requiring fully paired training data. 
\item \textbf{Benchmark \& Evaluation:} We establish a rigorous evaluation protocol by introducing the DF-RICO benchmark and a classifier-based metric to quantitatively measure the success of disentanglement and composition of imaging factors.
\end{itemize}

%% file: sec/2_related_works.tex
\section{Related Work}
\label{sec:related_work}

\subsubsection{Adapting Diffusion-Based Text-to-Image Models:}
Adapting pre-trained diffusion models enables personalization and domain-specific generation without re-training from scratch. Efficient fine-tuning methods such as HyperNetworks~\cite{ha2016hypernetworks}, LoRA~\cite{hu2022lora}, and DreamBooth~\cite{ruiz2023dreambooth} modify only a small subset of parameters in diffusion models for personalization. Spatial control methods like ControlNets~\cite{zhang2023adding} introduce additional conditions such as edges, depth, or segmentation. An extensive analysis of personalization techniques is given in~\cite{wei2025personalized}. Our proposed method differs from the above with the goal of controlling imaging factors such as the lens type, the sensor with its associated color profile, the viewpoint, and the domain.

\vspace{-0.5cm}
\subsubsection{Textual Inversion:}
Textual Inversion (TI)~\cite{gal2022image} introduces personalized concepts to T2I models by learning new token embeddings while keeping model weights frozen. The trainable concept token is optimized on a few examples containing the concept, enabling personalized image generation in a data- and parameter-efficient manner. The idea of TI has been applied to domains such as medical imaging~\cite{de2024medical} and style manipulations~\cite{butt2024colorpeel,sohn2023styledrop,xu2024dreamanime}. In some research, multiple class tokens are learned to maintain generation and classification capabilities~\cite{dong2025dreamartist,wang2025multi}. TI has also been applied to Content Retrieval~\cite{agnolucci2025isearle}, Person Re-Identification~\cite{baisa2025clip,yang2024pedestrian}, and Video Personalization~\cite{kansy2025reenact}. Our proposed method adapts TI for T2I diffusion models specifically to disentangle imaging factors.

\vspace{-0.5cm}
\subsubsection{Disentanglement of Concepts:}
Previous works explored the disentanglement of image content like visible objects and abstract concepts, such as color profiles and textures, in T2I diffusion models. Some improve expressiveness and compositionality by learning multiple tokens per image for separating distinct concepts~\cite{avrahami2023break,motamed2024lego}. Other works ~\cite{vinker2023concept,wei2023elite,xu2025cusconcept} introduce a hierarchical structure to the embeddings for the disentanglement of concepts. Some approaches utilize attention masks for disentanglement~\cite{shentu2024attencraft,zhang2024attention}. TokenVerse~\cite{garibi2025tokenverse} and Mod-Adapter~\cite{zhong2025mod} study unsupervised concept disentanglement by utilizing the modulation space of a transformer-based diffusion model. Our proposed method differs from the above in that we specifically focus on disentangling imaging factors.

%% file: sec/3_problem.tex
\section{Preliminaries}
\label{sec:preliminaries}

We formalize the problem of \textbf{Imaging Factor Disentanglement} as decomposing the image formation process into independent, physically meaningful primitives. Let $\mathcal{X} \subseteq \mathbb{R}^{H \times W \times 3}$ denote the space of 3-channel images with height $H$ and width $W$. We assume that the visual appearance of any image $x \in \mathcal{X}$ is governed by distinct imaging factor categories denoted with $k \in \mathcal{K}$, including lens, sensor, viewpoint and domain. For each category $k$, let $\mathcal{F}^{(k)}$ denote the space of all possible discrete values of that factor, for example $\mathcal{F}^{(\text{lens})} = \{\text{fisheye}, \text{rectilinear}, \dots\}$. The overall \textbf{factor space} is defined as the cartesian product $\mathcal{F} = \mathcal{F}^{(1)} \times \mathcal{F}^{(2)} \times \dots \times \mathcal{F}^{(K)}$. A specific imaging configuration is represented by a tuple $\mathbf{f} = (f^{(k)})_{k \in \mathcal{K}} \in \mathcal{F}$, where each $f^{(k)} \in \mathcal{F}^{(k)}$.

In real-world settings, it is typically infeasible to observe all possible combinations of imaging factors. Instead, we operate in a \textbf{sparse data regime}. We are given a collection of $N$ source datasets $\mathcal{D} = \{D^{(1)}, \dots, D^{(N)}\}$. Each dataset $D^{(i)}$ consists of $|D^{(i)}|$ image–factor pairs and is written as $D^{(i)} = \{(x^{(i)}_j, \mathbf{f}^{(i)}_j)\}_{j=1}^{|D^{(i)}|}$. Crucially, the joint distribution of factors within a given dataset $D^{(i)}$ is often fixed or highly coupled. For example, a dataset $D^{(1)}$ may exclusively contain images from a single domain captured with a specific sensor $s_A$, domain $d_A$, and lens $l_A$ while the viewpoint $v_A$ may change from image to image, such that \mbox{$\mathbf{f}^{(1)}_j = (s_A, l_A, d_A, v_A)$} $\forall j$. As a result, the union of observed factor configurations \[\mathcal{F}_{\text{obs}}
=\bigcup_{D \in \mathcal{D}}
\{\mathbf{f} \mid (x,\mathbf{f}) \in D\}
\subset \mathcal{F}
\] spans only a sparse subset of the full factor space $\mathcal{F}$.

Let $\mathcal{G}(\cdot, \theta_{\text{frozen}})$ denote a pre-trained generative model, such as a T2I diffusion model, with a set of frozen parameters $\theta_\text{frozen}$. We introduce a learned conditioning function $\psi: \mathcal{F} \rightarrow \mathbb{R}^{L \times d}$ that maps a factor tuple $\mathbf{f}$ to a sequence embedding of length $L$ and of dimension $d$. The objective is to learn $\psi$ such that samples $x \sim \mathcal{G}(\psi(\mathbf{f}))$ faithfully reflect the imaging characteristics specified by $\mathbf{f}$, that variations in any single factor $f^{(k)}$ are independent of all other factors $f^{(m)}$ for $m \neq k$, and that the model enables combinatorial generalization by synthesizing valid images for novel factor tuples $\mathbf{f}_{\text{new}} \in \mathcal{F}$ that are never observed in any of the source datasets $D\in\mathcal{D}$.

%% file: sec/4_method.tex
{}\begin{figure}[t]
    \centering
    \includegraphics[width=1\linewidth]{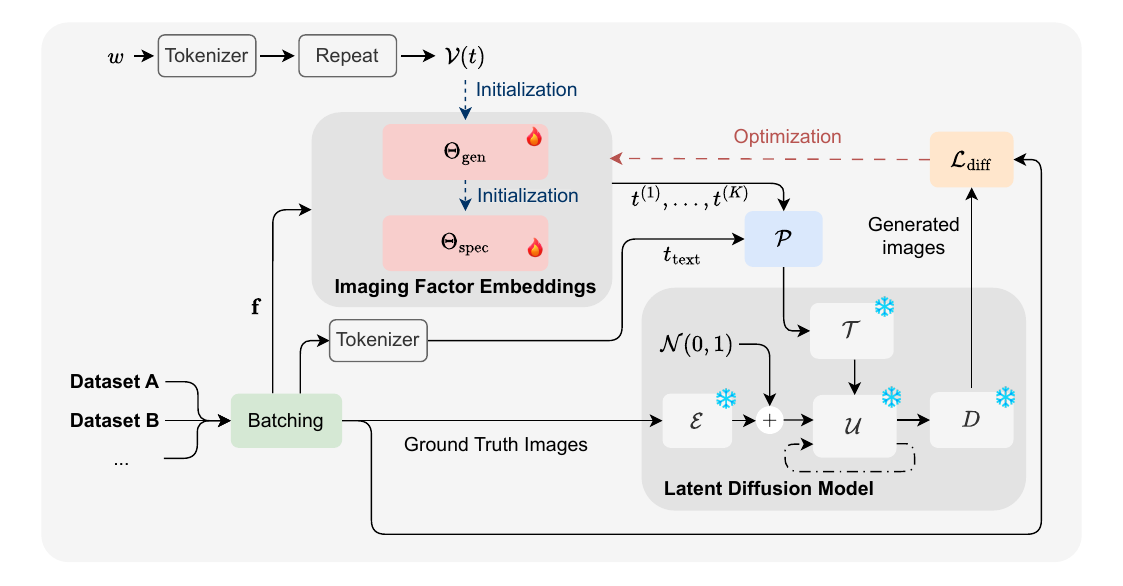}
    \caption{\textbf{Overview of the MULTI framework.} We optimize factor embeddings in a two-stage process: first, obtaining general embeddings, then refining them into dataset-specific ones. A specialized batching strategy is used to enforce factor overlap within each batch.}
    \label{fig:MULTI_framework}
    \vspace{-0.3cm}
\end{figure}

\section{Methodology}
\label{sec:method}

To address the problem of Imaging Factor Disentanglement defined in Sec.~\ref{sec:preliminaries}, we introduce \textbf{Mul}ti-factor disentanglement through \textbf{T}extual \textbf{I}nversion (\textbf{MULTI}). 
We utilize a Latent Diffusion Model (LDM) backbone consisting of a Variational Autoencoder (VAE) with encoder $\mathcal{E}$ and decoder $\mathcal{S}$ parameterized by weights $\theta_\mathcal{E}$ and $\theta_\mathcal{S}$, a text encoder $\mathcal{T}$ parameterized by $\theta_\mathcal{T}$, and a denoising U-Net $\mathcal{U}$ parameterized by $\theta_{\mathcal{U}}$. During training, an image $x$ is encoded into a latent code $z_0 = \mathcal{E}(x)$. Gaussian noise $\epsilon \sim \mathcal{N}(0, I)$ is added with a fixed variance schedule to produce a noisy latent $z_t$ at timestep $t$. The conditioning vector $\mathbf{c}$ is obtained by encoding a factor-based text prompt. The U-Net is trained to predict the noise $\epsilon$ conditioned on $t$ and $\mathbf{c}$:
\begin{equation}
    \mathcal{L}_{\mathrm{diff}} = \mathbb{E}_{z_t, x, \epsilon, t} \left[ || \epsilon - \mathcal{U}(z_t, t, \mathbf{c}) ||_2^2 \right]
    \label{eq:ldm_loss}
\end{equation}
In our framework, the parameters $\theta_\mathcal{E}, \theta_\mathcal{S}, \theta_\mathcal{T},$ and $\theta_\mathcal{U}$ remain frozen. 

\subsection{Factor Tokenization and Initialization}
\label{subsec:factor_tokenization_and_initialization}
We represent each factor value $f \in \mathcal{F}^{(k)}$ with a unique learnable token $t^{(k)}$. To capture the complexity of imaging factors (e.g., complex lens distortion profiles), we do not map a token to a single vector, but to a sequence $\mathcal{V}$ of $n$ learnable vectors $\mathcal{V}(t^{(k)}) = \{v_1^{(k)}, \dots, v_n^{(k)}\} \in \mathbb{R}^{n \times d}$, with $n$ being a hyperparameter of our method that controls the representational capacity allocated to each factor. 

\noindent \textbf{Initialization:} We initialize the learnable vectors $v \in \mathcal{V}$ using the semantic prior of the pre-trained text encoder to facilitate convergence. For a factor described by a word $w$ (e.g., ``fisheye''), we extract its embeddings $e_w \in \mathbb{R}^d$. For $n \textgreater 1$, we repeat $e_w$: $v_j^{(k)} = e_w$ for $j=1 \dots n$. This provides a semantically relevant starting point for optimization. Given the factor tuple $\mathbf{f} = (f^{(k)})_{k\in\mathcal{K}}$ of a training sample, we construct a conditioning factor-based text prompt $\mathcal{P}$. It is constructed by concatenating the learnable tokens for each factor with a tokenized natural language description of the image content $t_{\mathrm{text}}$:
\begin{equation}
    \mathcal{P}_{(\mathrm{S1},\mathbf{f})} = [t^{(1)}, t^{(2)}, \dots, t^{(K)}, t_{\mathrm{text}}]
    \label{eq:prompt_construction}
\end{equation}
This prompt is processed by text encoder $\mathcal{T}$ to produce the conditioning signal $\mathbf{c} = \mathcal{T}(\mathcal{P_{(\mathrm{S1},\mathbf{f})}})$. The embeddings corresponding to $t_\mathrm{text}$ is denoted by $\theta_{t_{\mathrm{text}}}$, which remains frozen. We employ a two-stage training strategy that separates the learning of general, dataset-independent imaging factors in the first stage and dataset-specific imaging factors in the second stage. 

\subsection{Stage-1: General Disentangled Factor Learning}
\label{subsec:stage_1}
The goal of the first stage is to learn a general, dataset-independent, disentangled representation for every factor value in the factor space $\mathcal{F}$. We denote the general factors as $f_{\mathrm{gen}}^{(k)} \in \mathcal{F}$. We optimize the set of all factor embeddings $\Theta_{\mathrm{gen}} = \{\mathcal{V}(t) \mid t \in \bigcup_k \mathcal{F}^{(k)}\}$. We train on the union of all available datasets $\mathcal{D}$. The optimization objective is to minimize the diffusion loss over all samples:
\begin{equation}
    \hat{\Theta}_{\mathrm{gen}} = 
    \argmin_{\Theta_{\mathrm{gen}}} 
    \mathbb{E}_{(x,\mathbf{f_{\mathrm{gen}}}) \sim \mathcal{D}}
    \left[ \mathcal{L}_{\mathrm{diff}}(x, \mathbf{f_{\mathrm{gen}}}; \Theta_{\mathrm{gen}}) \right]
    \label{eq:stage1_opt}
\end{equation}
The expectation is approximated using mini-batches during training. A critical challenge is preventing the model from learning spurious correlations between factors that frequently co-occur in specific datasets. To mitigate this, we  explicitly assemble training batches to include samples that share a specific factor $f^{(k)}$ (e.g., ``fisheye'') while the remaining factors as well as their source datasets (e.g., $D^{(a)}$ and $D^{(b)}$) differ. By minimizing the loss across these varied contexts simultaneously, the factor embedding $\mathcal{V}(t^{(k)})$ is compelled to capture only the common visual characteristics, effectively decoupling it from dataset-specific confounds. This batching strategy reduces cross-factor leakage from correlated factors in DF-RICO, although complete conditional independence cannot be guaranteed.

\subsection{Stage-2: Dataset-Specific Factor Adaptation}
\label{subsec:stage_2}
Real-world datasets often contain unique sensor noise or domain shifts that can not be fully captured by the general factors $f_\mathrm{gen}^{(k)} \in \mathcal{F}$. To address this, during the second stage we optimize specific factors tied to a dataset $D^{(i)}$.

For each factor token $t^{(k)}$ relevant to dataset $D^{(i)}$, we initialize a specific token $t_\mathrm{spec}^{(k, i)}$ using the learned weights from Stage-1. Let $\Theta_{\mathrm{spec}}$ denote the set of these specific embeddings. To prevent the model from overfitting or entangling the factor embeddings, we employ a stochastic token masking strategy.

During a training step for dataset $D^{(i)}$, we select a random factor category $k^* \sim \mathrm{Uniform}(\mathcal{K})$. The corresponding factor tuple is updated to $\mathbf{f} = (f_{\mathrm{gen}}^{(1)}, \dots, f_{\mathrm{spec}}^{(k^*)}, \dots, f_{\mathrm{gen}}^{(K)})$, while keeping other factor tokens fixed to their Stage-1 states. The prompt becomes $\mathcal{P}_{(\mathrm{S2},\textbf{f})} = [t_{\mathrm{gen}}^{(1)}, \dots, t_{\mathrm{spec}}^{(k^*, i)}, \dots, t_{\mathrm{gen}}^{(K)}, C_{\mathrm{text}}]$ The gradient updates are applied only to the specific embedding $\mathcal{V}(t_{\mathrm{spec}}^{(k^*, i)})$. This encourages the specific token to capture residual domain-specific information (e.g., exact sensor noise) for category $k^*$ in dataset $D^{(i)}$, without entangling factors.

\subsection{Inference and Combination}
\label{subsec:inference}
After training, we can generate novel images by freely combining factors. To synthesize a scene with factor tuple 
$\mathbf{f}_{\mathrm{novel}} = (f_{\mathrm{novel}}^{(1)}, \dots, f_{\mathrm{novel}}^{(K)})$, where 
$f_{\mathrm{novel}}^{(k)} \in \{f_{\mathrm{gen}}^{(k)},  {f_{\mathrm{spec}}}^{(k,D)}\}$
and $D \in \mathcal{D}$. We construct the prompt using the corresponding tokens $\mathcal{P}_{\mathrm{novel}}$. This allows for the combinatorial generalization required to fill the missing regions of the sparse factor space $\mathcal{F}$.
We can also generate novel images by conditioning our model with $\mathcal{P}_{\mathrm{novel}}$ and an edge or depth control map~\cite{zhang2023adding}.

%% file: sec/5_experiments.tex
\newcolumntype{M}[1]{>{\centering\arraybackslash}m{#1}}
\newcolumntype{M}[1]{>{\centering\arraybackslash}m{#1}}

\begin{table}[t]
\caption{\textbf{Qualitative comparison of existing combinations.}
The first row shows a sample image from the actual dataset. DreamBooth and MULTI perform on a similar level and reach mostly images close to the reference image.}
\centering
\scriptsize
\setlength{\tabcolsep}{2pt}
\renewcommand{\arraystretch}{1.3}

\makebox[\textwidth][c]{%
\resizebox{1.0\textwidth}{!}{%
\begin{tabular}{l *{10}{>{\centering\arraybackslash}m{2.2cm}}}

\textbf{Domain} &
\textbf{Real} &
\textbf{Real} &
\textbf{Real} &
\textbf{Real} &
\textbf{Real} &
\textbf{Real} &
\textbf{Real} &
\textbf{Real} &
\textbf{Simulation} &
\textbf{Video Game} \\

\textbf{Sensor} &
\textbf{RGB} &
\textbf{RGB-Thermal} &
\textbf{Gated} &
\textbf{Event} &
\textbf{Thermal} &
\textbf{RGB} &
\textbf{RGB} &
\textbf{Thermal} &
\textbf{RGB} &
\textbf{RGB} \\

\textbf{Viewpoint} &
\textbf{Front} &
\textbf{Front} &
\textbf{Front} &
\textbf{Front} &
\textbf{Front} &
\textbf{Back} &
\textbf{Drone} &
\textbf{Pole} &
\textbf{Front} &
\textbf{Front} \\

\textbf{Lens} &
\textbf{Normal} &
\textbf{Normal} &
\textbf{Normal} &
\textbf{Normal} &
\textbf{Normal} &
\textbf{Fisheye} &
\textbf{Fisheye} &
\textbf{Fisheye} &
\textbf{Normal} &
\textbf{Normal} \\

\textbf{Original}
& \roundedimg{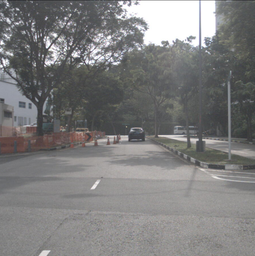}
& \roundedimg{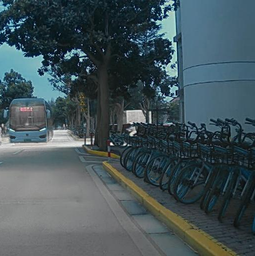}
& \roundedimg{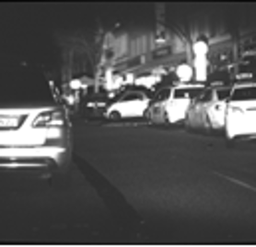}
& \roundedimg{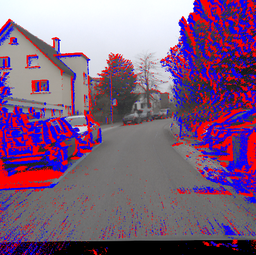}
& \roundedimg{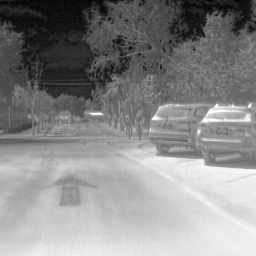}
& \roundedimg{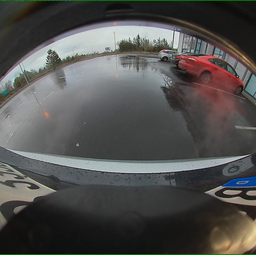}
& \roundedimg{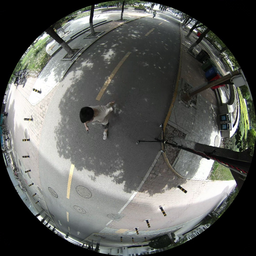}
& \roundedimg{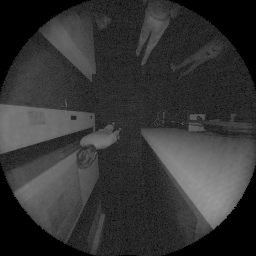}
& \roundedimg{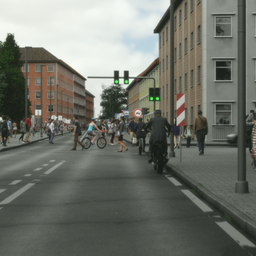}
& \roundedimg{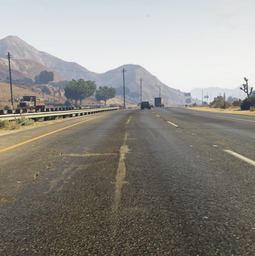} \\

\textbf{Zeroshot~\cite{podell2023sdxl}}
& \roundedimg{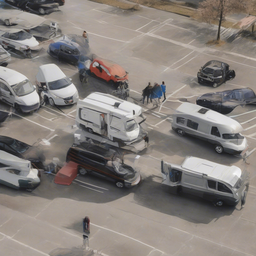}
& \roundedimg{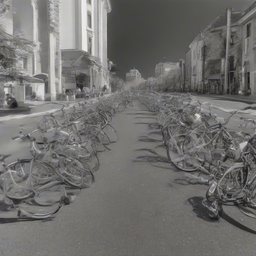}
& \roundedimg{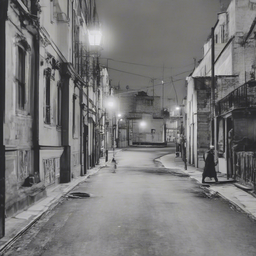}
& \roundedimg{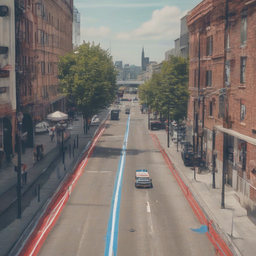}
& \roundedimg{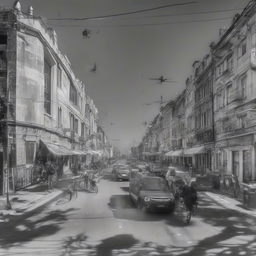}
& \roundedimg{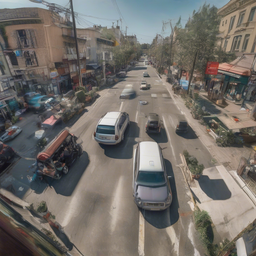}
& \roundedimg{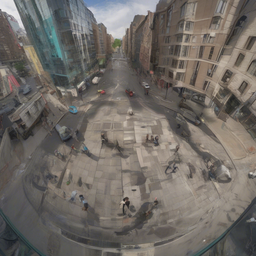}
& \roundedimg{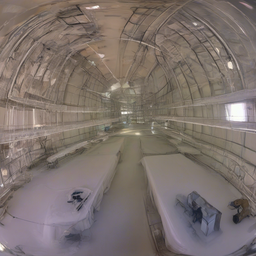}
& \roundedimg{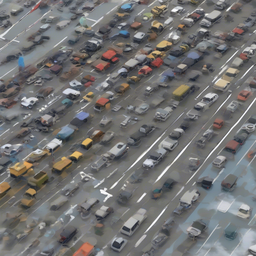}
& \roundedimg{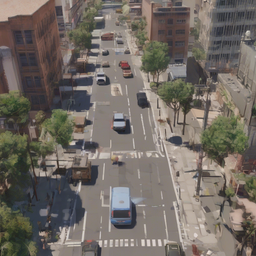} \\

\textbf{DreamBooth~\cite{ruiz2023dreambooth}}
& \roundedimg{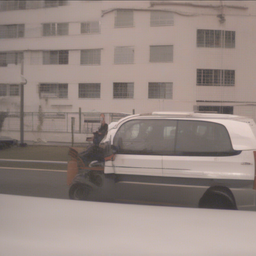}
& \roundedimg{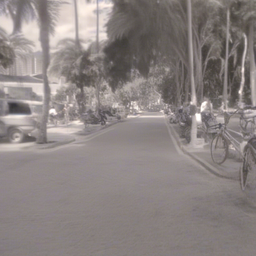}
& \roundedimg{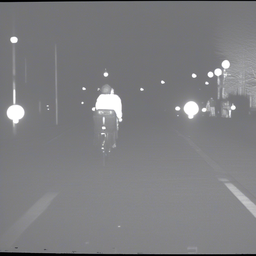}
& \roundedimg{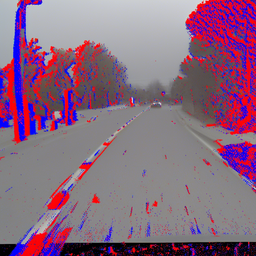}
& \roundedimg{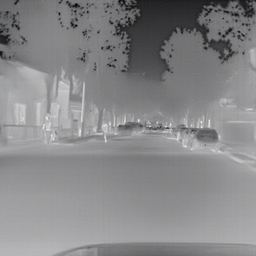}
& \roundedimg{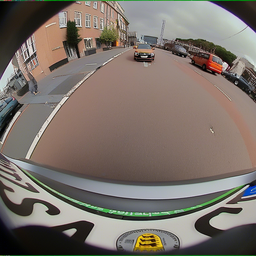}
& \roundedimg{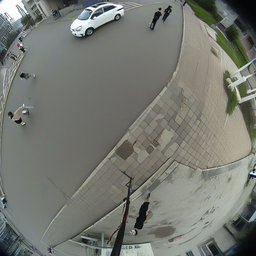}
& \roundedimg{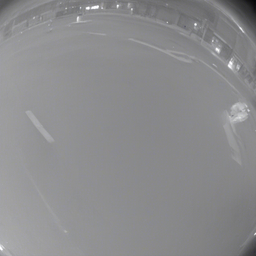}
& \roundedimg{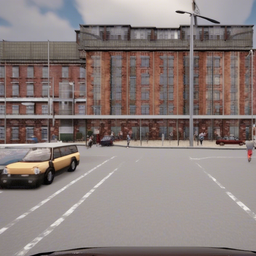}
& \roundedimg{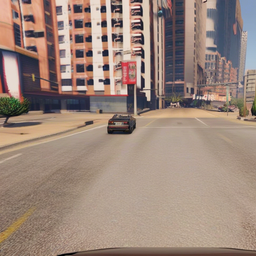} \\

\textbf{Inspiration Tree~\cite{vinker2023concept}}
& \roundedimg{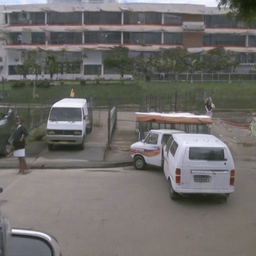}
& \roundedimg{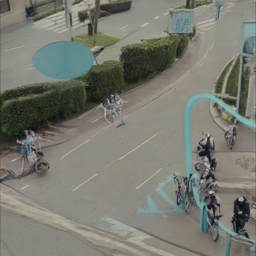}
& \roundedimg{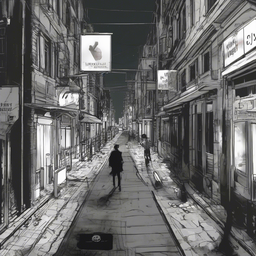}
& \roundedimg{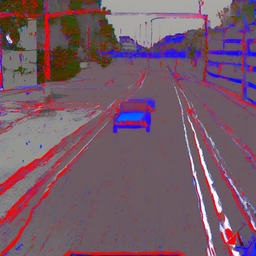}
& \roundedimg{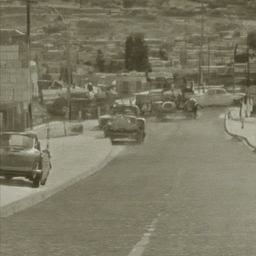}
& \roundedimg{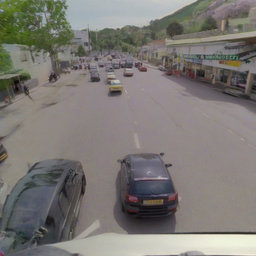}
& \roundedimg{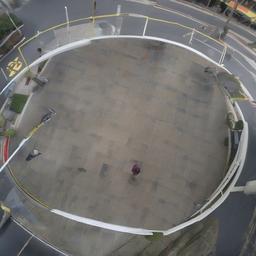}
& \roundedimg{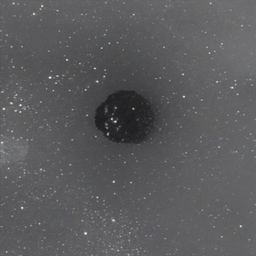}
& \roundedimg{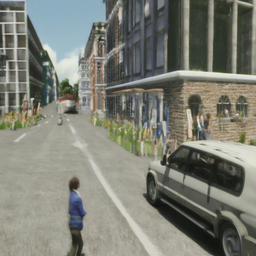}
& \roundedimg{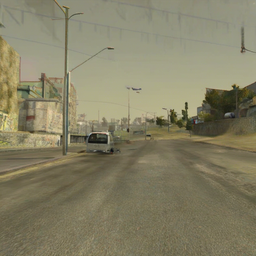} \\

\textbf{MULTI (Ours)}
& \roundedimg{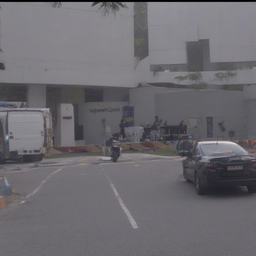}
& \roundedimg{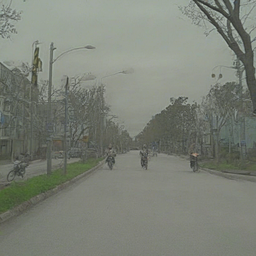}
& \roundedimg{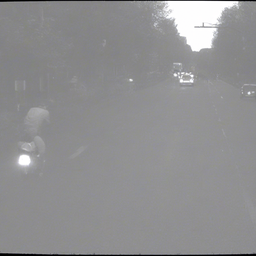}
& \roundedimg{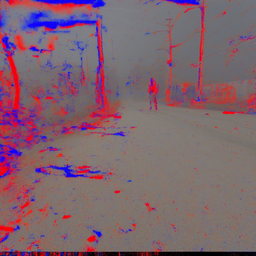}
& \roundedimg{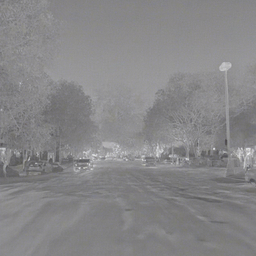}
& \roundedimg{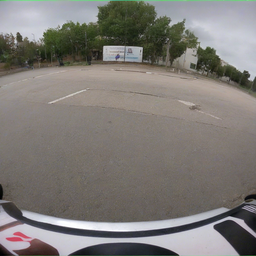}
& \roundedimg{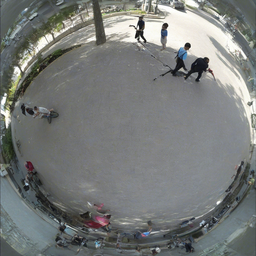}
& \roundedimg{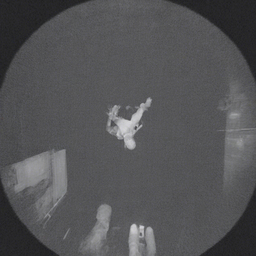}
& \roundedimg{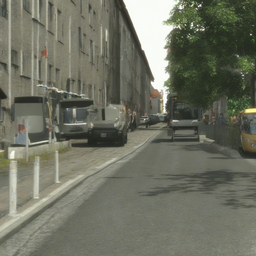}
& \roundedimg{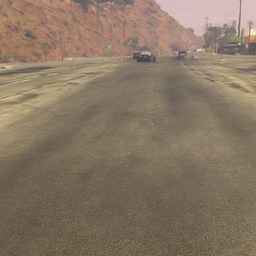} \\

\end{tabular}%
}}
\label{tab:qualitative_existing_combinations}
\end{table}

\begin{table}[t]
\caption{\textbf{Qualitative comparison of novel factor combinations.} 
Novel combinations are unseen during training. While both DreamBooth and MULTI generalize across simultaneous changes in source, sensor, view, and lens, MULTI demonstrates marginally superior robustness.}

\centering
\small
\setlength{\tabcolsep}{2pt}
\renewcommand{\arraystretch}{1.15}

\makebox[\textwidth][c]{%
\resizebox{1.0\textwidth}{!}{%
\begin{tabular}{l *{10}{>{\centering\arraybackslash}m{2.2cm}}}
\textbf{Domain} &
\textbf{Real} &
\textbf{Simulation} &
\textbf{Real} &
\textbf{Real} &
\makecell{\textbf{Simulation}\\(shift)} &
\makecell{\textbf{Simulation}\\(synscape)} &
\makecell{\textbf{Video Game}\\(sim10k)} &
\textbf{Simulation} &
\textbf{Real} &
\textbf{Real} \\

\textbf{Sensor} &
\makecell{\textbf{Thermal}\\(teledyne)} &
\textbf{Thermal} &
\makecell{\textbf{Gated}\\(dense)} &
\makecell{\textbf{Event}\\(dsec)} &
\makecell{\textbf{Thermal}\\(timo)} &
\textbf{RGB} &
\makecell{\textbf{Gated}\\(dense)} &
\textbf{Event} &
\textbf{Thermal} &
\textbf{RGB-Thermal} \\

\textbf{Viewpoint} &
\textbf{Drone} &
\textbf{Front} &
\textbf{Pole} &
\makecell{\textbf{Drone}\\(visdrone)} &
\textbf{Front} &
\makecell{\textbf{Drone}\\(loaf)} &
\makecell{\textbf{Drone}\\(loaf)} &
\textbf{Front} &
\textbf{Pole} &
\textbf{Front} \\

\textbf{Lens} &
\textbf{Fisheye} &
\makecell{\textbf{Fisheye}\\(woodscape)} &
\textbf{Fisheye} &
\textbf{Normal} &
\textbf{Normal} &
\textbf{Normal} &
\textbf{Normal} &
\textbf{Fisheye} &
\textbf{Normal} &
\textbf{Fisheye} \\

        \textbf{Zeroshot~\cite{podell2023sdxl}}
        & \roundedimg{images/examples/NovelComb_SDZS_1.png}
        & \roundedimg{images/examples/NovelComb_SDZS_2.png}
        & \roundedimg{images/examples/NovelComb_SDZS_3.png}
        & \roundedimg{images/examples/NovelComb_SDZS_4.png}
        & \roundedimg{images/examples/NovelComb_SDZS_5.png}
        & \roundedimg{images/examples/NovelComb_SDZS_6.png}
        & \roundedimg{images/examples/NovelComb_SDZS_7.png}
        & \roundedimg{images/examples/NovelComb_SDZS_8.png}
        & \roundedimg{images/examples/NovelComb_SDZS_9.png}
        & \roundedimg{images/examples/NovelComb_SDZS_10.png}\\

        \textbf{DreamBooth~\cite{ruiz2023dreambooth}}
        & \roundedimg{images/examples/NovelComb_DB_1.png}
        & \roundedimg{images/examples/NovelComb_DB_2.png}
        & \roundedimg{images/examples/NovelComb_DB_3.png}
        & \roundedimg{images/examples/NovelComb_DB_4.png}
        & \roundedimg{images/examples/NovelComb_DB_5.png}
        & \roundedimg{images/examples/NovelComb_DB_6.png}
        & \roundedimg{images/examples/NovelComb_DB_7.png} 
        & \roundedimg{images/examples/NovelComb_DB_8.png}
        & \roundedimg{images/examples/NovelComb_DB_9.png}
        & \roundedimg{images/examples/NovelComb_DB_10.png} \\

        \textbf{MULTI (Ours)}
        & \roundedimg{images/examples/NovelComb_MULTI_1.png}
        & \roundedimg{images/examples/NovelComb_MULTI_2.png}
        & \roundedimg{images/examples/NovelComb_MULTI_3.png}
        & \roundedimg{images/examples/NovelComb_MULTI_4.png}
        & \roundedimg{images/examples/NovelComb_MULTI_5.png}
        & \roundedimg{images/examples/NovelComb_MULTI_6.png}
        & \roundedimg{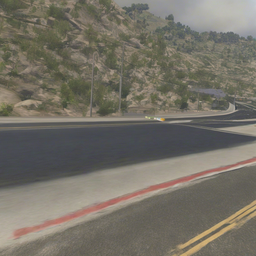}
        & \roundedimg{images/examples/NovelComb_MULTI_8.png}
        & \roundedimg{images/examples/NovelComb_MULTI_9.png}
        & \roundedimg{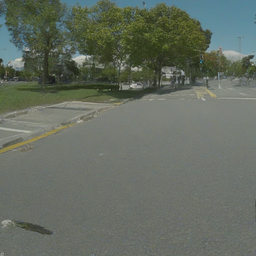}\\

        \end{tabular}%
    }
}
\label{tab:qualitative_novel_combinations}
\end{table}

\begin{table}[t]
\caption{\textbf{Qualitative comparison of novel factor combinations with ControlNets.}
Image-to-image generation using ControlNet guidance with altered factors unseen during training.}
\centering
\footnotesize
\setlength{\tabcolsep}{2pt}
\renewcommand{\arraystretch}{1.3}

\makebox[\textwidth][c]{%
\resizebox{1.0\textwidth}{!}{%
\begin{tabular}{l *{10}{>{\centering\arraybackslash}m{2.2cm}}}

\textbf{Domain} &
\textbf{Real} &
\textbf{Video Game} &
\textbf{Real} &
\textbf{Real} &
\textbf{Real} &
\textbf{Real} &
\textbf{Video Game} &
\textbf{Real} &
\textbf{Simulation} &
\textbf{Real} \\

\textbf{Sensor} &
\textbf{RGB} &
\textbf{RGB} &
\textbf{RGB} &
\textbf{RGB} &
\textbf{Event} &
\textbf{RGB} &
\textbf{RGB} &
\textbf{Thermal} &
\textbf{RGB} &
\textbf{RGB-Thermal} \\

\textbf{Viewpoint} &
\textbf{Front} &
\textbf{Front} &
\textbf{Front} &
\textbf{Front} &
\textbf{Front} &
\textbf{Drone} &
\textbf{Front} &
\textbf{Front} &
\textbf{Front} &
\textbf{Back} \\

\textbf{Lens} &
\textbf{Normal} &
\textbf{Normal} &
\textbf{Normal} &
\textbf{Normal} &
\textbf{Normal} &
\textbf{Fisheye} &
\textbf{Normal} &
\textbf{Normal} &
\textbf{Normal} &
\textbf{Fisheye} \\

\textbf{Original} &
\roundedimg{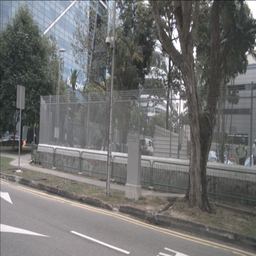} &
\roundedimg{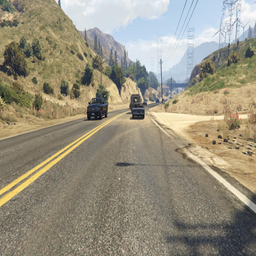} &
\roundedimg{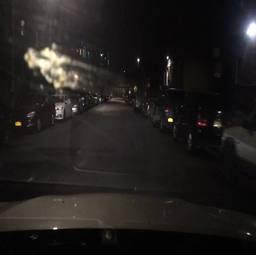} &
\roundedimg{} &
\roundedimg{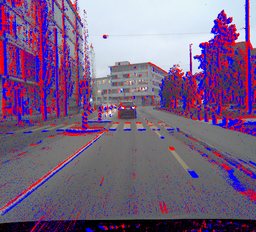} &
\roundedimg{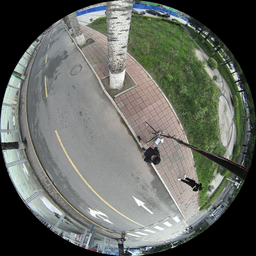} &
\roundedimg{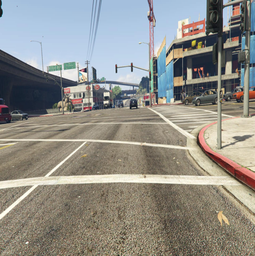} &
\roundedimg{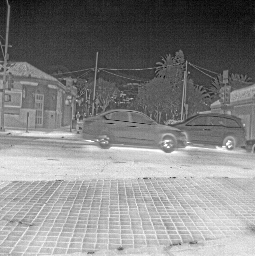} &
\roundedimg{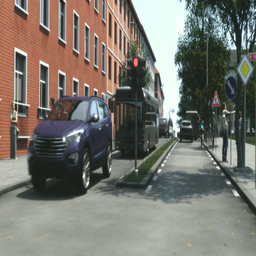} &
\roundedimg{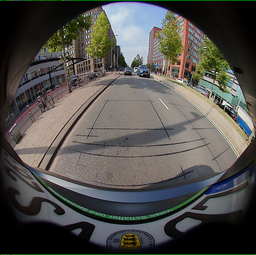} \\

\textbf{ControlNet Input} &
\roundedimg{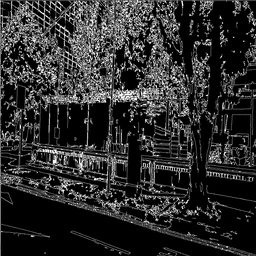} &
\roundedimg{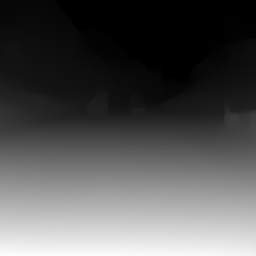} &
\roundedimg{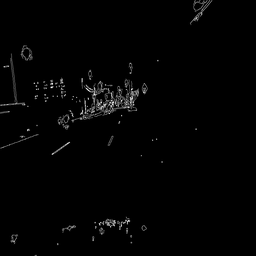} &
\roundedimg{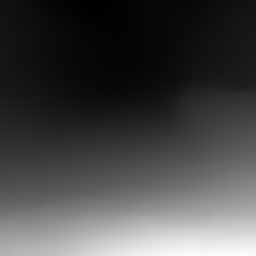} &
\roundedimg{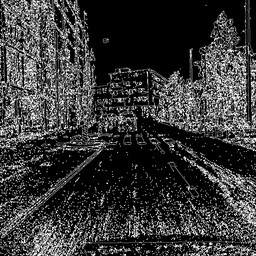} &
\roundedimg{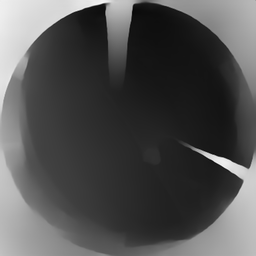} &
\roundedimg{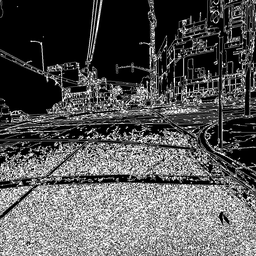} &
\roundedimg{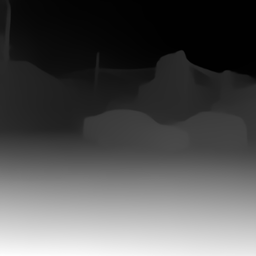} &
\roundedimg{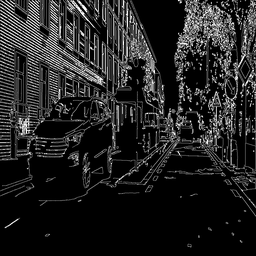} &
\roundedimg{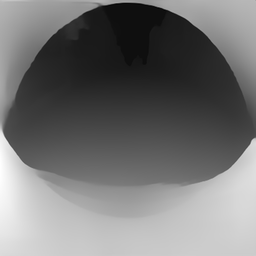} \\

\textbf{Altered Factor} &
\makecell{\textbf{Real}\\ (nuimages)\\$\downarrow$\\ \textbf{Video Game}\\ (sim10k)} &
\makecell{\textbf{RGB}\\$\downarrow$\\ \textbf{Thermal}\\ (teledyne)} &
\makecell{\textbf{RGB}\\$\downarrow$\\ \textbf{Gated}\\ (dense)} &
\makecell{\textbf{RGB}\\ (bdd100k)\\$\downarrow$\\ \textbf{Event}\\ (dsec)} &
\makecell{\textbf{Real}\\$\downarrow$\\ \textbf{Simulation}\\ (shift)} &
\makecell{\textbf{Real}\\$\downarrow$\\ \textbf{Video Game}\\ (sim10k)} &
\makecell{\textbf{Video Game}\\ (sim10k)\\$\downarrow$\\ \textbf{Real}\\ (nuimages)} &
\makecell{\textbf{Real}\\$\downarrow$\\ \textbf{Simulation}\\ (shift)} &
\makecell{\textbf{RGB}\\$\downarrow$\\ \textbf{Thermal}\\ (teledyne)} &
\makecell{\textbf{RGB}\\$\downarrow$\\ \textbf{RGB-Thermal}} \\

\textbf{Zeroshot~\cite{podell2023sdxl}} &
\roundedimg{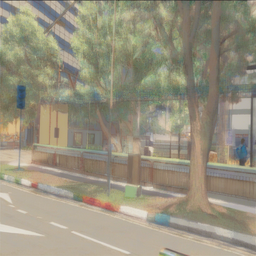} &
\roundedimg{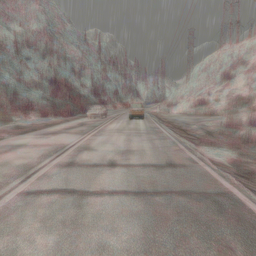} &
\roundedimg{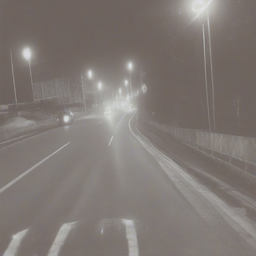} &
\roundedimg{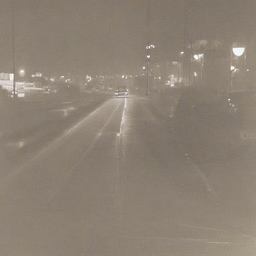} &
\roundedimg{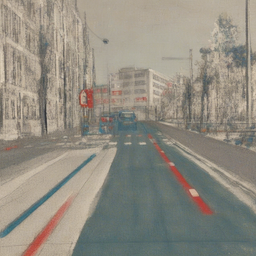} &
\roundedimg{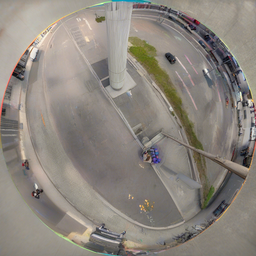} &
\roundedimg{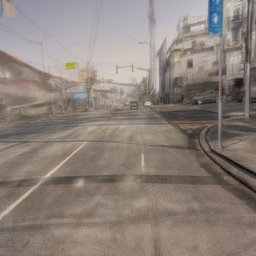} &
\roundedimg{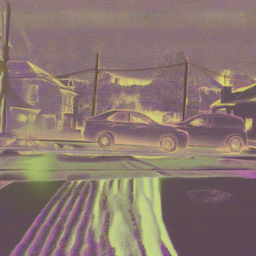} &
\roundedimg{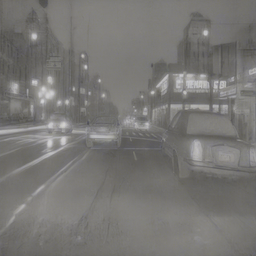} &
\roundedimg{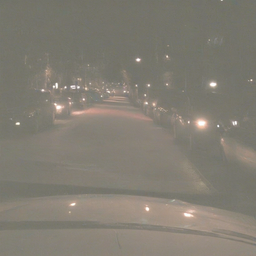} \\

\textbf{DreamBooth~\cite{ruiz2023dreambooth}} &
\roundedimg{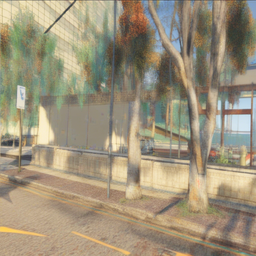} &
\roundedimg{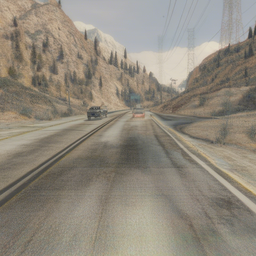} &
\roundedimg{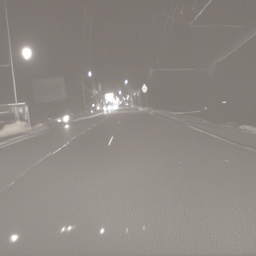} &
\roundedimg{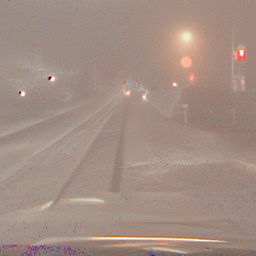} &
\roundedimg{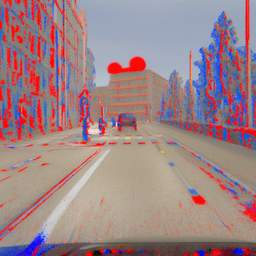} &
\roundedimg{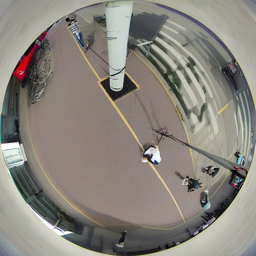} &
\roundedimg{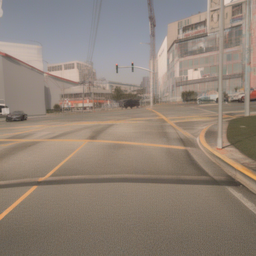} &
\roundedimg{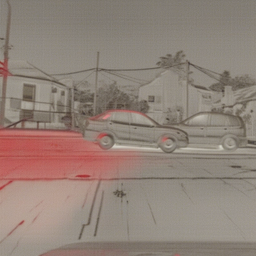} &
\roundedimg{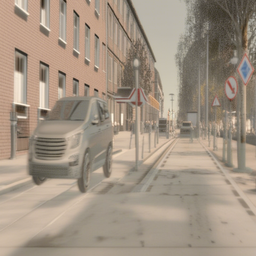} &
\roundedimg{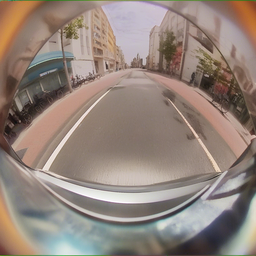} \\

\textbf{MULTI (Ours)} &
\roundedimg{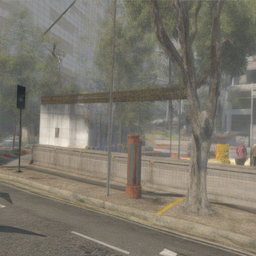} &
\roundedimg{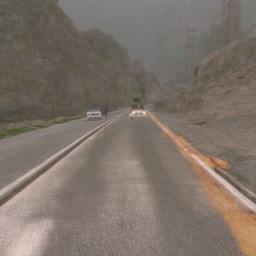} &
\roundedimg{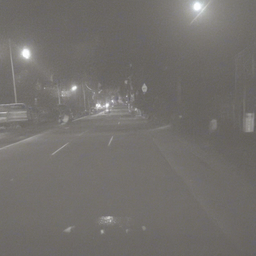} &
\roundedimg{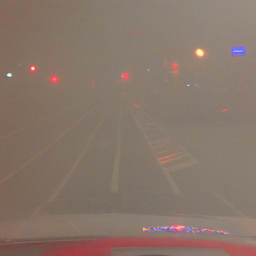} &
\roundedimg{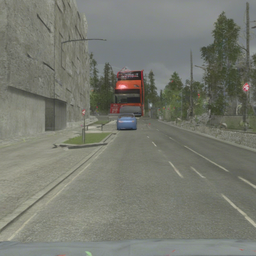} &
\roundedimg{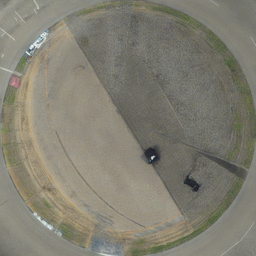} &
\roundedimg{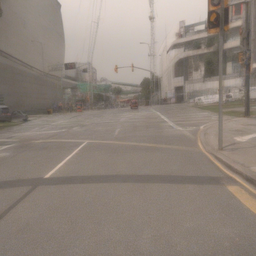} &
\roundedimg{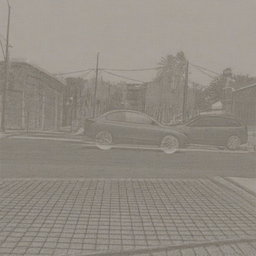} &
\roundedimg{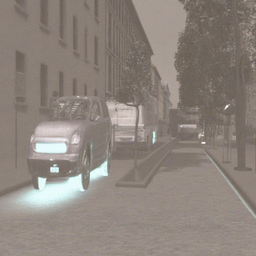} &
\roundedimg{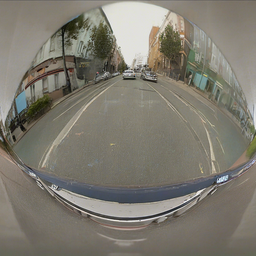} \\

\end{tabular}%
}%
}%
\label{tab:qualitative_controlnets_novel_combinations}
\end{table}

\begin{table*}[t]
\caption{\textbf{User study comparison.} Results across controllability, existing, and novel combinations. Best in bold.}
\label{tab:user_study}
\centering
\scriptsize
\renewcommand{\arraystretch}{1.05}
\setlength{\tabcolsep}{6pt}

\begin{tabular}{lccc}
\toprule
\textbf{Method} 
& \textbf{Controllability $\uparrow$ [\%]} 
& \textbf{Existing $\uparrow$ [\%]} 
& \textbf{Novel  $\uparrow$[\%]} \\
\midrule
Zeroshot~\cite{podell2023sdxl}             & 21.16 &  6.22  & 14.11   \\
DreamBooth~\cite{ruiz2023dreambooth}      & 27.69  & 45.03  & 32.37  \\
MULTI (Ours)      & \textbf{51.15 } & \textbf{48.75  } & \textbf{53.53 } \\
\bottomrule
\end{tabular}
\end{table*}

\begin{table*}[t]
\caption{\textbf{Comparison on generating existing factor combinations.} The factor combinations resemble the existing ones from the training dataset. Best in bold.}
\label{table:quantitaive_existing_factors}
\centering
\scriptsize
\renewcommand{\arraystretch}{1.05}
\setlength{\tabcolsep}{2.2pt}
\begin{tabular}{l cccc cccc c}
\toprule
\textbf{Method}
& \multicolumn{4}{c}{\textbf{Generation Metrics}}
& \multicolumn{5}{c}{\textbf{Factor-Alignment Accuracy}} \\
\cmidrule(lr){2-5} \cmidrule(lr){6-10}
& FID $\downarrow$ & CLIP $\uparrow$ & IS $\uparrow$ & DS $\uparrow$
& Sensor $\uparrow$ & Lens $\uparrow$ & View $\uparrow$ & Domain $\uparrow$ & \textbf{Avg.} $\uparrow$
\\
\midrule

Zeroshot~\cite{podell2023sdxl}
& 60.19 & 25.13 & 4.36 & 0.59
& 0.81 & 0.71 & 0.29 & 0.79 & 0.65 \\

DreamBooth~\cite{ruiz2023dreambooth}
& 68.27 & 24.86 & 5.02 & 0.57
& \textbf{0.95} & \textbf{0.94} & \textbf{0.76 }& \textbf{0.96} & \textbf{0.90} \\

Inspiration Tree~\cite{vinker2023concept}
& 62.35 & 23.77 & \textbf{5.08} & 0.58
&  0.78 & 0.68 & 0.34 & 0.71 & 0.63 \\

MULTI (ours)
& \textbf{60.13} & \textbf{27.30} & 4.68 & \textbf{0.65}
& 0.94 & \textbf{0.94} & 0.74 & 0.92 & 0.88 \\

\midrule

Zeroshot-Canny
& 53.85 & \textbf{31.69} & 4.62 & \textbf{0.58}
& 0.87 & 0.96 & 0.72 & 0.94 &  0.87\\

DreamBooth-Canny
& 70.98 & 27.17 & 4.65 & 0.56
& 0.96 & \textbf{0.99} & 0.87 & 0.98 &  0.95\\

MULTI-canny (ours)
& \textbf{45.39} & 28.33 & \textbf{4.69} & 0.55
& \textbf{0.97} & \textbf{0.99} & \textbf{0.91} & \textbf{0.99} & \textbf{0.97} \\

\midrule

Zeroshot-Depth
& 62.38 & \textbf{31.22} & 4.75 & \textbf{0.58}
& 0.87 & 0.96 & 0.73 & 0.91 &  0.87\\

DreamBooth-Depth
& 69.59 & 28.27 & 4.53 & 0.56
& \textbf{0.95} & \textbf{0.97} & 0.86 & 0.98 &  0.94\\

MULTI-depth (ours)
& \textbf{60.60} & 27.93 & \textbf{4.78} & \textbf{0.55}
& \textbf{0.95} & \textbf{0.97} & \textbf{0.91} & \textbf{0.99} & \textbf{0.96} \\
\bottomrule
\end{tabular}

\end{table*}

\begin{table*}[t]
\caption{\textbf{Comparison on generating novel factor combinations.} The result consists of novel combinations from lens, sensor, viewpoint and domain that are not present in the training datasets. Best in bold.}
\label{table:quantitaive_new_factors}
\centering
\scriptsize
\renewcommand{\arraystretch}{1.05}
\setlength{\tabcolsep}{2.2pt}
\begin{tabular}{l ccc cccc c}
\toprule
\textbf{Method}
& \multicolumn{3}{c}{\textbf{Generation Metrics}}
& \multicolumn{5}{c}{\textbf{Factor-Alignment Accuracy}} \\
\cmidrule(lr){2-4} \cmidrule(lr){5-9}
 & CLIP $\uparrow$ & IS $\uparrow$ & DS $\uparrow$
& Sensor $\uparrow$ & Lens $\uparrow$ & View $\uparrow$ & Domain $\uparrow$ & \textbf{Avg.} $\uparrow$
\\
\midrule

Zeroshot~\cite{podell2023sdxl}
& 22.01 & 1.23 & 0.53 & 0.06 & \textbf{0.43} & \textbf{0.53} & \textbf{0.56} & 0.40  \\

DreamBooth~\cite{ruiz2023dreambooth}
& 27.13 & \textbf{1.86 } & 0.53 & 0.39 & 0.39 & 0.33  & \textbf{0.56} & 0.42  \\


MULTI (ours)
& \textbf{27.32} & 1.79 & \textbf{0.57} & \textbf{0.66} & 0.39 & 0.13 & \textbf{0.56} & \textbf{ 0.44}  \\

\midrule

Zeroshot-Canny
& \textbf{27.64} & 5.73 & 0.51 & 0.81 & 0.95 & 0.71 & 0.91 & 0.85  \\

DreamBooth-Canny
& 27.12 & \textbf{5.77} & 0.50 & \textbf{0.82} & \textbf{0.99} & 0.81 & \textbf{0.91} & \textbf{0.88}  \\

MULTI-canny (ours)
& 26.72 & 4.88 & \textbf{0.58} & \textbf{0.82} & \textbf{0.99} & \textbf{0.83} & 0.89 & \textbf{0.88}  \\

\midrule

Zeroshot-Depth
& 27.82 & 5.48 & 0.52 & 0.81 & 0.94 & 0.73 & 0.89 & 0.84 \\

DreamBooth-Depth
& \textbf{27.94} & \textbf{5.56} & 0.52 & \textbf{0.82} & 0.94 & 0.78 & \textbf{0.90} & \textbf{0.86}  \\

MULTI-depth (ours)
& 26.82 & 4.36 & \textbf{0.54} & 0.80 & \textbf{0.98} & \textbf{0.80} & 0.94 & \textbf{0.86}  \\

\bottomrule
\end{tabular}
\vspace{-0.3cm}
\end{table*}

\section{Experimental Results}
\label{sec:exp_results}
\subsubsection{Implementation Details:}  We use Stable Diffusion XL (SDXL)~\cite{podell2023sdxl} as the backbone model. We utilize both of the text encoders of the SDXL model to learn separate factor embeddings. We employ the AdamW optimizer~\cite{Loshchilov2017DecoupledWD} with weight decay of $10^{-2}$, warmup, and cosine learning rate decay with a maximum learning rate of $10^{-4}$. We set the number of learnable vectors to $n=15$ corresponding to $15 \times (768+1024)$ trainable parameters per factor embedding. We train for 10 epochs with a batch size of $4$ for both stages. We use $25$ diffusion steps for all generated images and a guidance scale of $2.5$.

\vspace{-0.5cm}
\subsubsection{Evaluation:}
We evaluate our method using standard image generation metrics: Fréchet Inception Distance (FID)~\cite{heusel2017gans} to measure distributional similarity between real and generated images, CLIP Score~\cite{hessel2021clipscore} to assess semantic alignment with factor-based text prompts, Inception Score (IS)~\cite{salimans2016improved} to evaluate image quality, and Diversity Score (DS)~\cite{zhang2018unreasonable} to quantify perceptual diversity. To measure disentanglement and compositional generalization, we introduce \textit{Factor-Alignment Accuracy} (FAA), which classifies the presence of each factor in generated images using the DINOv3~\cite{simeoni2025dinov3} feature extractor and a classifier composed of four linear layers trained on factor labels. We additionally conduct a user study to assess human preference and perceived factor fidelity of generated images.

All images are defined by combinations of factors, enabling evaluation across two scenarios: existing combinations present in the training datasets, and novel combinations to assess recomposition of disentangled factors into unseen configurations. For each combination, we evaluate two settings: T2I, where images are generated from factor-based text prompts, and I2I, where images are generated conditionally using factor-based text prompts together with edge or depth maps for structural guidance, with a single factor altered to form novel combinations.

\vspace{-0.5cm}
\subsubsection{Dataset and setup:}
We employ datasets from the RICO benchmark~\cite{neuwirth2025rico}, which consists of $15$ datasets for autonomous driving and surveillance applications. Although it was initially developed for domain-incremental learning in object detection, the datasets provide diverse imaging factor configurations, making them suitable for our investigations. In this work, we introduce a new benchmark called Disentangling Factors with RICO (DF-RICO), where each dataset is explicitly characterized by combination of imaging factors, enabling systematic evaluation of factor disentanglement and novel factor recombination in generative models.

We define $4$ factor categories $k$: camera lens, sensor, viewpoint, and domain. Each category is associated with a set of possible values (1) camera lens: normal, fisheye; (2) sensor: rgb, thermal, rgb-thermal, gated, event; (3) viewpoint: front, back, side, drone, pole; (4) domain: real, video-game, simulation. For each dataset, we label the corresponding factors from each factor category and generate structured prompts, e.g., "A <source\_real>  <sensor\_rgb> image from <view\_front> captured with <lens\_fisheye> showing ....", and the description of the image is generated using the BLIP model~\cite{li2022blip}, typically containing 12-15 tokens. 

\vspace{-0.5cm}
\subsubsection{Baseline:} We compare our method against three baselines: (1) SDXL Zeroshot~\cite{podell2023sdxl} which is the unadapted diffusion-based model; (2) DreamBooth~\cite{ruiz2023dreambooth}, a fine-tuning-based personalization technique that trains the entire diffusion model; (3) Inspiration Tree~\cite{vinker2023concept}, a hierarchical, binary-tree structure for disentanglement of stylistic components.

\subsection{Results}
\label{subsec:results}
\subsubsection{Qualitative Results:}
In Table~\ref{tab:qualitative_existing_combinations}, we depict the results of images generated using existing combinations of imaging factors across the baselines and our proposed method. 
MULTI, Inspiration Tree, and DreamBooth generate visually appealing images under these setting, whereas Zeroshot fails to align with the visual distribution of our dataset. MULTI produces images that more closely align with the target dataset characteristics, exhibiting consistent dataset-specific color grading and closely matches the distortion of a fisheye lens. Moreover, for rgb-thermal images, MULTI reflects the rgb-thermal characteristics rather than collapsing to purely thermal appearances. 

In Table~\ref{tab:qualitative_novel_combinations}, we present qualitative results of images using novel combinations of imaging factors comparing the baselines and our proposed method. Inspiration Tree is excluded as a baseline for novel combinations because it decomposes a concept token into two sub tokens, and these sub tokens do not correspond the interpretable imaging factors. Zeroshot fails to depict any of the intended factors accurately, whereas both MULTI and DreamBooth show partial factor adherence. DreamBooth tends to emphasize a narrow subset of factors (lens type and viewpoint), whereas MULTI better preserves a broader subset of factors (viewpoint, domain, and sensor). Novel image synthesis requires maintaining multiple independent factors simultaneously, indicating that MULTI more effectively disentangles and recombines factors. 

In Table~\ref{tab:qualitative_controlnets_novel_combinations}, we show I2I results under novel factor combinations. Inspiration Tree is excluded from this setting for the same reasons discussed earlier. All the methods successfully preserve the scene layout provided by edge or depth maps. Zeroshot lacks the visual fidelity of the target dataset. While DreamBooth and MULTI achieve comparable image quality, MULTI more cleanly reflects the altered factor, while keeping the non-altered factors consistent.

\vspace{-0.5cm}
\subsubsection{Quantitative Results:}
We conducted a user study with 27 participants who evaluated unlabeled and randomly ordered images generated by different methods across three categories: existing combinations, novel combinations, and ControlNet guided generation. In Table~\ref{tab:user_study}, we show preference scores across controllability (structure preservation), existing combination, and novel combinations. Users prefer MULTI in all aspects, with particularly strong gains for novel combinations. 

In Table~\ref{table:quantitaive_existing_factors}, we depict image generation metrics for existing factor combinations with and without utilizing ControlNets. While our method exhibits the best FID score, it performs worse with other image generation metrics. However, it attains high FAA, indicating that MULTI generates images that reflect the intended factor combinations specified in the prompt better than the other approaches. In Table~\ref{table:quantitaive_new_factors}, we report image quality metrics for novel factor combinations with and without utilizing ControlNets. While our method performs slightly lower on certain image quality metrics, it achieves the highest FAA.




\subsection{Ablation Study}
\label{subsec:ablation_studies}

\begin{figure}[t]
    \centering
    \begin{subfigure}[b]{0.25\textwidth}
        \includegraphics[height=2.4cm,keepaspectratio]{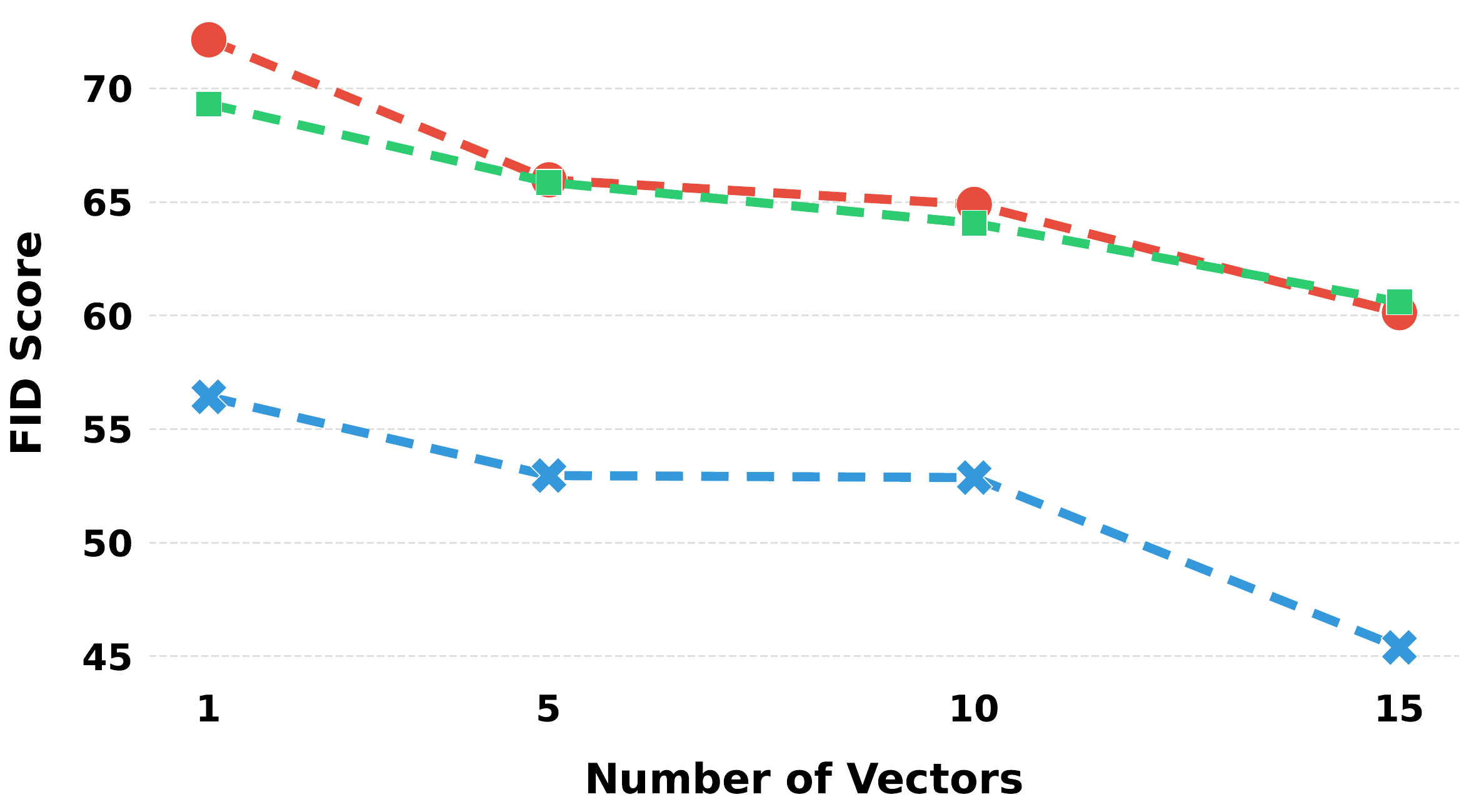}
    \end{subfigure}\hfill
    \begin{subfigure}[b]{0.25\textwidth}
        \includegraphics[height=2.4cm,keepaspectratio]{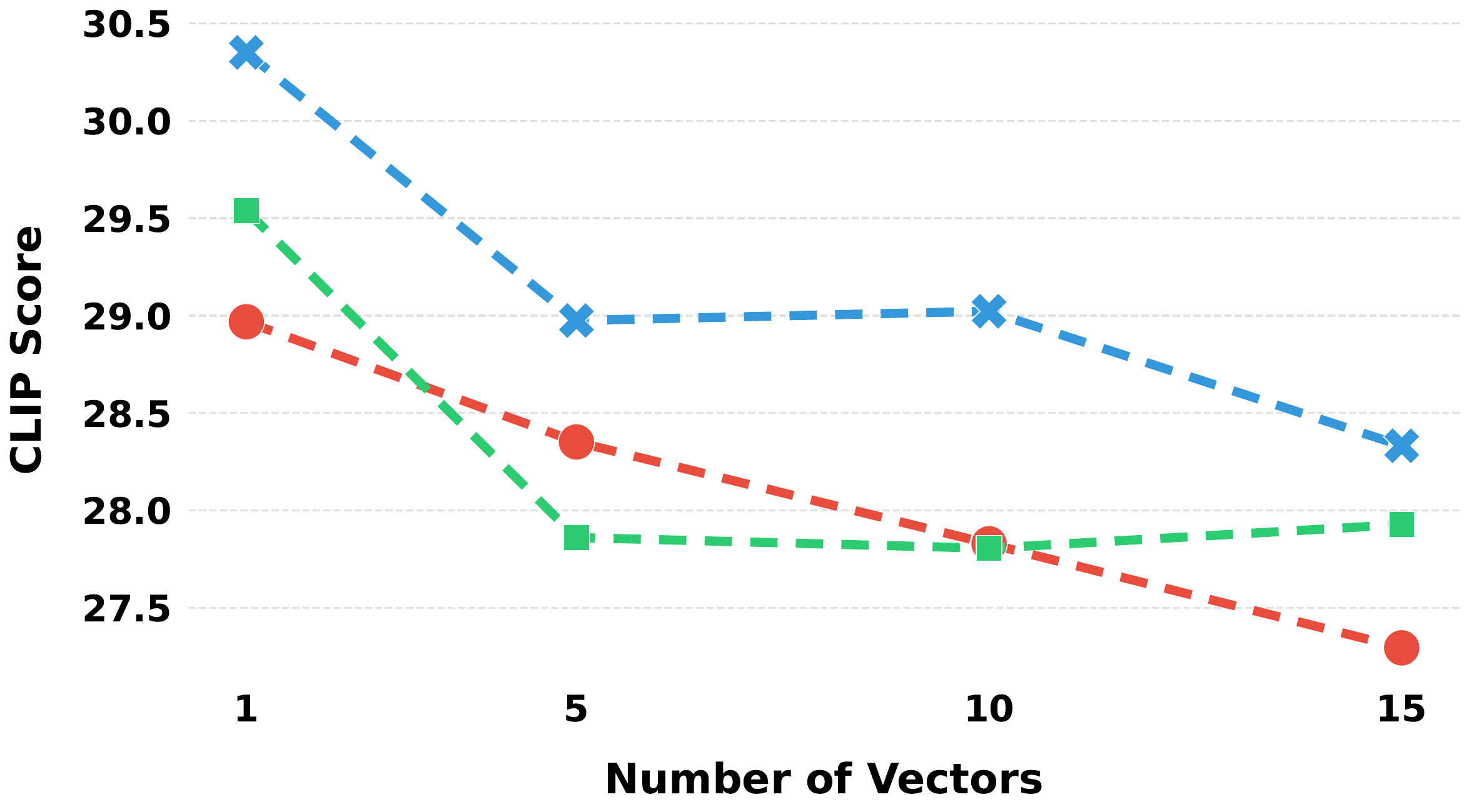}
    \end{subfigure}\hfill
    \begin{subfigure}[b]{0.25\textwidth}
        \includegraphics[height=2.4cm,keepaspectratio]{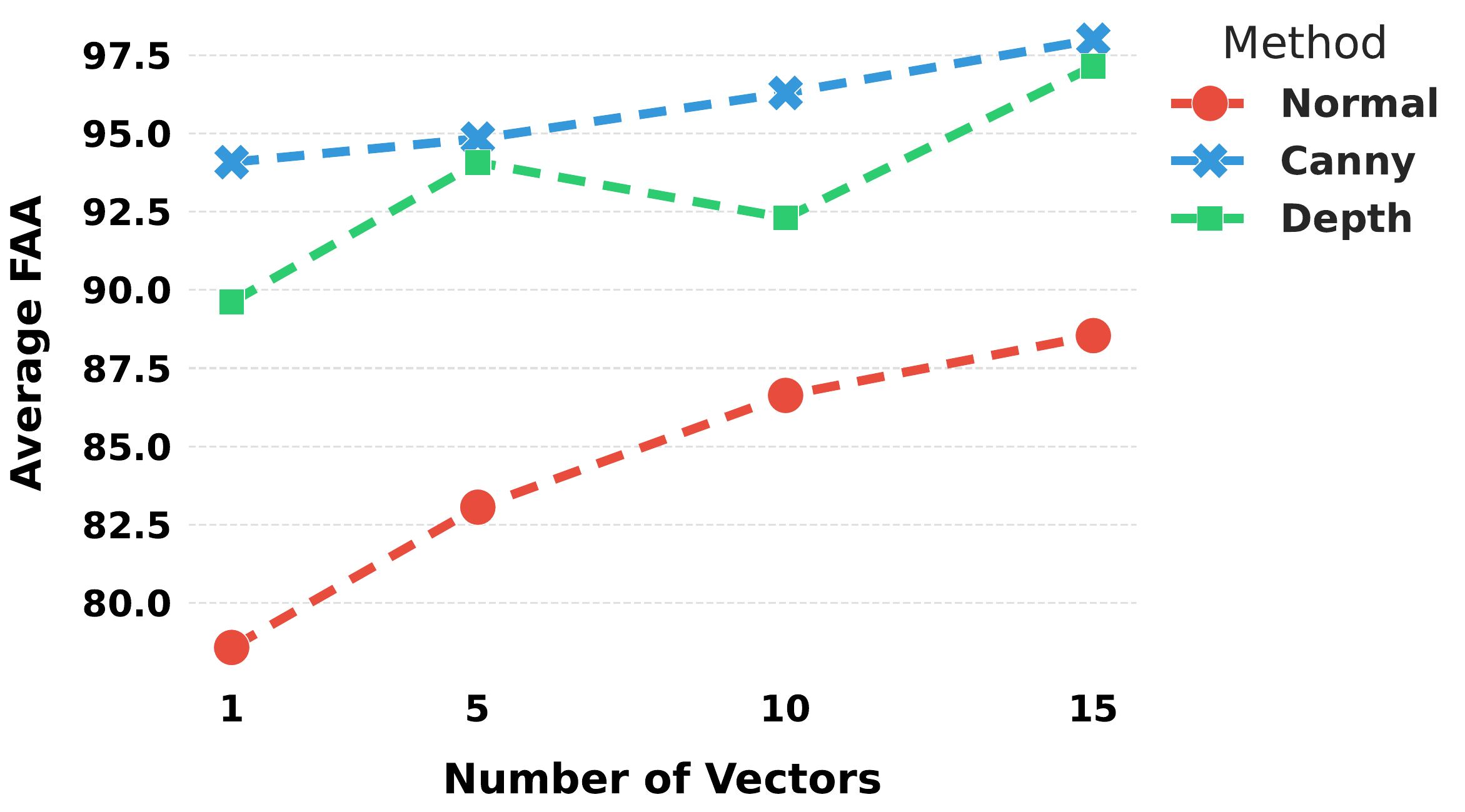}
    \end{subfigure}
    \caption{\textbf{Effect of the number of learnable vectors} and ControlNets on FID (left) and CLIP score (middle), and FAA (right).}
    \label{fig:affect_of_num_vecs_with_controlnets}
\end{figure}

\begin{figure}[t]
\centering
    \centering
    \begin{subfigure}[b]{0.45\textwidth}
        \centering
        \includegraphics[height=2.4cm,keepaspectratio]{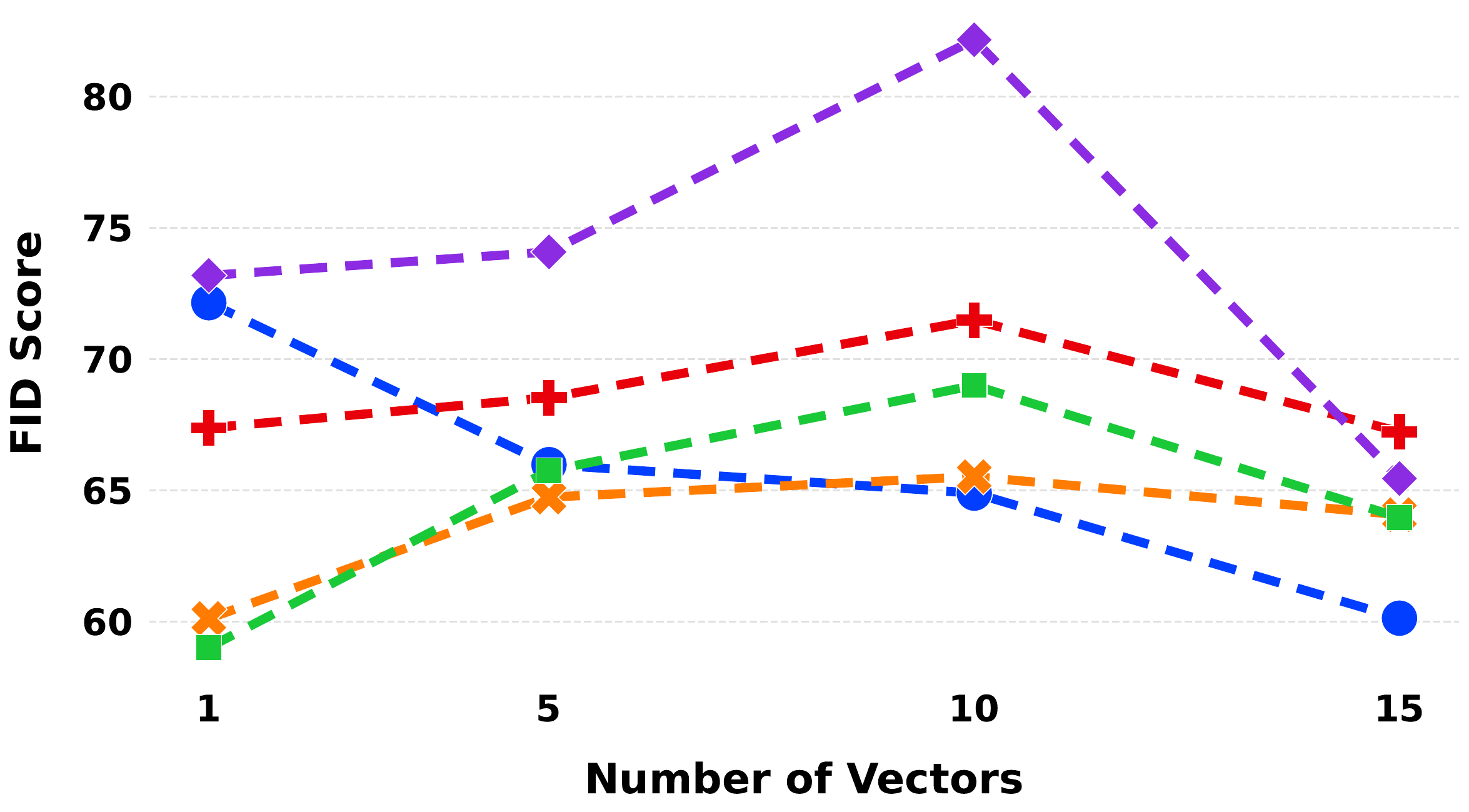}
    \end{subfigure}
    \hspace{0.05\textwidth}
    \begin{subfigure}[b]{0.45\textwidth}
        \centering
        \includegraphics[height=2.4cm,keepaspectratio]{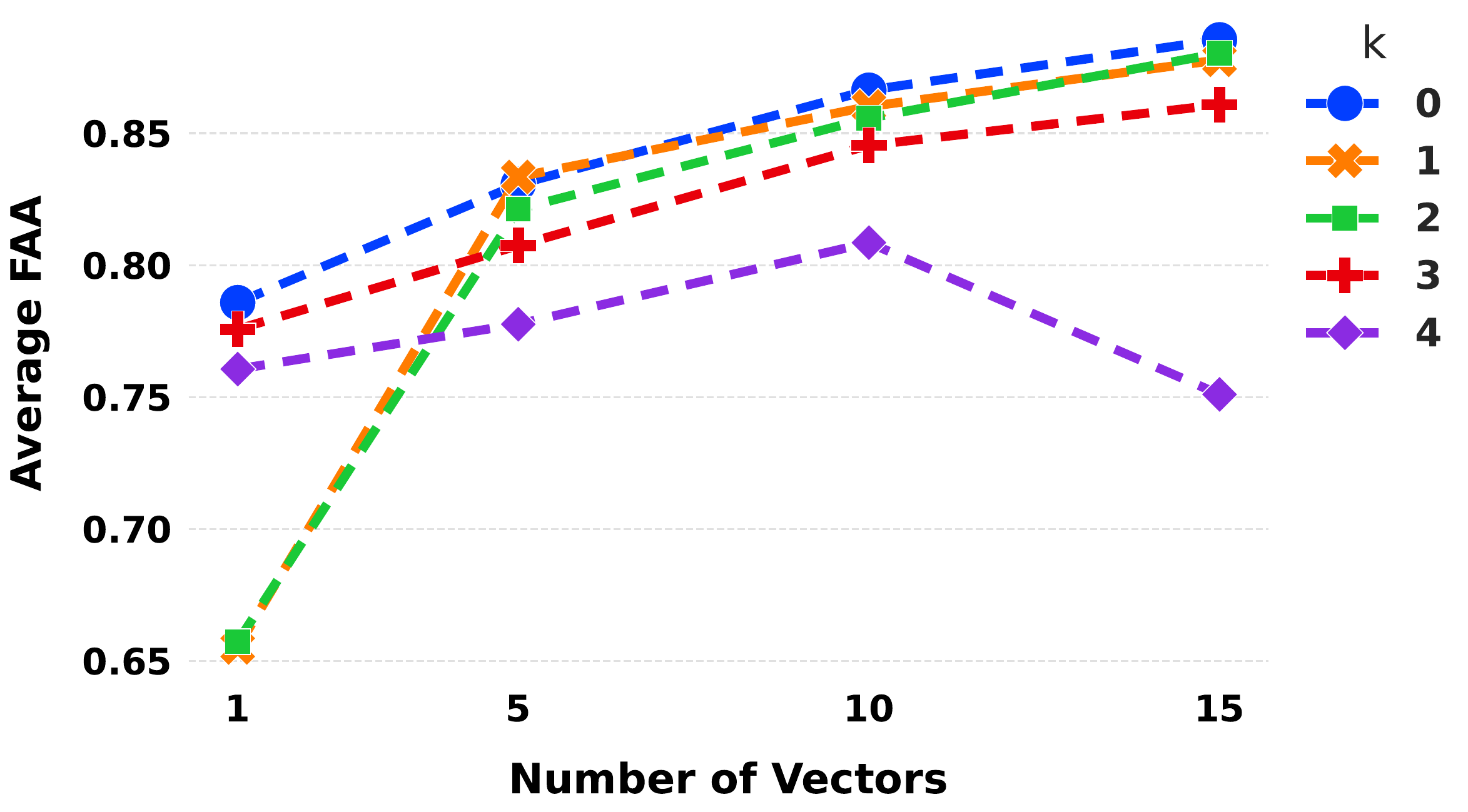}
    \end{subfigure}

    \caption{
    \textbf{Effect of Fraction of general and specific factors} in the prompt on FID and FAA. Here, $k$ denotes the number of general factor tokens in the prompt, while $k=4$
    corresponds to dataset-specific tokens.
    }
    \label{fig:affect_of_num_vecs_with_num_of_general_tokens}
    \vspace{-0.3cm}
\end{figure}

The number of learnable vectors per factor token controls the representational capacity of each factor. Figure~\ref{fig:affect_of_num_vecs_with_controlnets} analyzes the impact of the vector count $n$ and identifies $n=15$ as the optimal value, achieving both low FID and high FAA. As $n$ increases, the CLIP score decreases, likely since the factor embeddings overshadow the descriptive prompt. Larger values of $n$ are constrained by CLIP’s fixed token length, which would displace natural language embeddings required for semantic guidance. The use of ControlNets follows a similar trend.

In addition, the fraction of general versus specific factors in the prompt determines the balance between generalization and dataset-dependent specialization. Figure~\ref{fig:affect_of_num_vecs_with_num_of_general_tokens} shows that increasing the proportion of specific factors improves image quality and FAA, particularly for larger $n$, validating the effectiveness of the two-stage design.

%% file: sec/6_analysis.tex
\section{Discussion}
\label{sec:discussion}

\subsubsection{Regarding Novel Combinations:}
Although personalization enables T2I models to reproduce specific patterns of individual datasets, explicitly disentangling the influence of imaging factors remains a challenging problem. The baselines and our approach exhibit reduced performance on novel combinations of these factors. $\blacktriangleright$~\textit{Imaging-factor disentanglement remains hard.}

\vspace{-0.5cm}
\subsubsection{Regarding MULTI:}
Our method performs well on imaging factor combinations observed during training. Although it improves upon the baselines for novel combinations, the presence of artifacts indicates that imaging factor disentanglement remains a challenging problem, for which our method represents an initial step towards its resolution. $\blacktriangleright$~\textit{MULTI is an initial step towards imaging factor disentanglement.}

\vspace{-0.5cm}
\subsubsection{Regarding Evaluations:}
Evaluating novel imaging factor combinations lacks a reference distribution. Hence, we propose using a classifier to detect the presence of each factor. This allows to evaluate factors independently, but the classifier may not fully capture all relevant aspects of imaging factors. The training method and model architecture of the classifier also becomes an influential component of this evaluation strategy.
$\blacktriangleright$~\textit{Classifier-based evaluation supports analyzing novel factor combinations.}

\vspace{-0.5cm}
\subsubsection{Regarding ControlNets:}
Our results show that canny edge maps and depth maps improve performance for lens and viewpoint factors. These control modalities provide auxiliary spatial cues for image generation, thereby improving the performance for the imaging factors that strongly influence the spatial layout of the image. This holds true for MULTI, DreamBooth, and Zeroshot settings. We therefore report T2I and ControlNet-guided results separately. ControlNets also enable indirect image-to-image generation.
$\blacktriangleright$~\textit{ControlNets offer training-free spatial control for MULTI.}

%% file: sec/7_conclusion.tex
\section{Conclusion}
\label{sec:conclusion}
We are the first to introduce the problem of disentangling imaging factors to generate images with novel combinations thereof. We propose \textbf{MULTI}, a method designed to disentangle imaging factors and generalize to combinations beyond the configurations observed during training. Our experiments reveal that existing approaches for personalization, as well as content-style separation methods, struggle with this challenge, and that our approach improves upon this. Although performance remains somewhat inconsistent, our method improves factor disentanglement, leading to better performance for novel combinations. This indicates that imaging factor disentanglement remains a difficult challenge and an open research problem. This work serves as a foundation for future work, motivating the development of more sophisticated and specialized approaches to solve imaging factor disentanglement and controllable generation of novel factor combinations. 

%% file: supplementary.tex




\newcommand{\maketitlesupplementary}{%
  \clearpage
  \thispagestyle{plain}

  \vspace*{-\topskip}

  \begin{center}
    {\Large\bfseries \thetitle\par}
    \vspace{0.5em}
    {\Large\normalfont Supplementary Material\par} 
    \vspace{1.0em}
  \end{center}
}








\clearpage

\appendix

  \begin{center}
    \vspace{0.5em}
    {\Large\normalfont Supplementary Material\par} 
    \vspace{1.0em}
  \end{center}

\renewcommand{\thesection}{S.\arabic{section}}

\renewcommand{\thefigure}{S.\arabic{figure}}
\renewcommand{\thetable}{S.\arabic{table}}
\setcounter{figure}{0}
\setcounter{table}{0}

\section{DF-RICO Benchmark}

To study our method, we build upon RICO benchmark~\cite{neuwirth2025rico}, which was initially developed for incremental learning for object detection. We introduce DF-RICO (\textbf{D}isentangling \textbf{F}actors with \textbf{RICO}), which spans a wide range of datasets and each dataset is explicitly associated with specific configuration of imaging factors, including camera lens, sensor, viewpoint, and domain.
In MULTI, we define a structured prompt template: "A <source> <sensor> image from <view> captured with <lens> showing <description>.", where <source>, <sensor>, <view>, and <lens> correspond to learnable factor tokens that are shared across datasets, while <description> describes the image content. The content description is generated for each image using Bootstrapping Language–Image Pretraining (BLIP) model~\cite{li2022blip}, a vision language model that integrates image and text understanding. This enables consistent prompt construction without manual annotation.
This structured design enables controlled generation and evaluation of various factor combinations. Table~\ref{tab:dfrico_benchamrk} shows the datasets from DF-RICO benchmark and their associated factor combinations.

\begin{table}[h]
\caption{\textbf{DF-RICO Benchmark} with 15 datasets and its corresponding factors.}
\centering
\scriptsize
\begin{tabular}{l l c c c c}
\hline
 \textbf{Task Name} &\textbf{Dataset} & \textbf{Sensor} & \textbf{Viewpoint} & \textbf{Lens} & \textbf{Domain} \\
\hline
 \textit{daytime} &{nuImages}~\cite{caesar_nuscenes_2020} & RGB & Back/Front & Normal & Real \\
 \textit{thermal} &{Teledyne FLIR}~\cite{teledyne_flir_free_2024} & Thermal & Front & Normal & Real \\
 \textit{fisheye fix} &{Fisheye8K}~\cite{gochoo_fisheye8k_2023} & RGB & Pole & Fisheye & Real \\
 \textit{drone} &{VisDrone}~\cite{zhu_detection_2022} & RGB & Drone & Normal & Real \\
 \textit{simulation} &{SHIFT}~\cite{sun_shift_2022} & RGB & Front & Normal & Simulation \\
 \textit{fisheye car} &{WoodScape}~\cite{yogamani_woodscape_2021} & RGB & Back/Front/Side & Fisheye & Real \\
 \textit{RGB + thermal fusion} &{SMOD}~\cite{chen_amfd_2024} & RGB Thermal & Front & Normal & Real \\
 \textit{video game} &{Sim10K}~\cite{johnson-roberson_driving_2017} & RGB & Front & Normal & Video Game \\
 \textit{nighttime} &{BDD100K}~\cite{yu_bdd100k_2020} & RGB & Front & Normal & Real \\
 \textit{fisheye indoor} &{LOAF}~\cite{yang_large-scale_2023} & RGB & Drone & Fisheye & Real \\
 \textit{gated} &{DENSE}~\cite{bijelic_seeing_2020} & Gated & Front & Normal & Real \\
 \textit{photoreal. simulation} &{Synscapes}~\cite{wrenninge_synscapes_2018} & RGB & Front & Normal & Simulation \\
 \textit{thermal fisheye indoor} &{TIMo}~\cite{schneider_timodataset_2022} & Thermal & Pole & Fisheye & Real \\
 \textit{inclement} &{DENSE}~\cite{bijelic_seeing_2020} & RGB & Front & Normal & Real \\
 \textit{event camera} &{DSEC}~\cite{gehrig_low-latency_2024,gehrig_dsec_2021} & Event & Front & Normal & Real \\
\hline
\end{tabular}
\label{tab:dfrico_benchamrk}
\end{table}

\section{Factor Alignment Accuracy}
To evaluate the success of disentangling factors, we introduce Factor Alignment Accuracy (FAA), a new metric that builds upon an imaging factor classifier. FAA is computed using a multi-head image classifier that independently predicts the presence of each factor from a generated image.

\subsubsection{Classifier Architecture}
The classifier is built upon a pre-trained \mbox{DINOv3}~\cite{simeoni2025dinov3} backbone, which is kept frozen. We adapt pre-trained \mbox{DINOv3}~\cite{simeoni2025dinov3} model to prevent overfitting under data sparsity and to improve the generalization of novel factor combinations. For an input image, the backbone extracts the class token, i.e., global visual representation, which is shared across four independent linear classifier heads, with softmax normalization~\cite{goodfellow2016deep}. Each of these heads corresponds to one of the imaging factor categories: lens, sensor, viewpoint, and domain. Each head performs multi-class classification over the discrete set of values defined for its corresponding factor category.

\subsubsection{Training}
During training of the classifier, the backbone weights are kept frozen, and the four linear classification heads are trained. The training samples consist of real images paired with factor labels as defined in Table.~\ref{tab:dfrico_benchamrk}. The classifier is trained for $10$ epochs using AdamW optimizer~\cite{Loshchilov2017DecoupledWD} with weight decay of $10^{-2}$ and a learning rate of $10^{-4}$. The total loss is computed as the sum of cross-entropy losses~\cite{bishop2006pattern} across all four classifier heads. This encourages the classifier to focus on visually grounded factors.

\subsubsection{FAA computation for evaluation}
During evaluation, the trained classifier is applied to the generated images. For each generated image, the classifier predicts the factor labels independently for all four factor categories. FAA measures the fraction of images for which the predicted factor matches the specified factor. FAA measures semantic alignment between factors and images, making it suitable for assessing disentanglement.

\section{Imaging Factor Embedding Space Visualization}
\begin{figure}[h]
    \centering
    \begin{minipage}{0.48\textwidth}
        \centering
        \includegraphics[height=4cm,keepaspectratio]{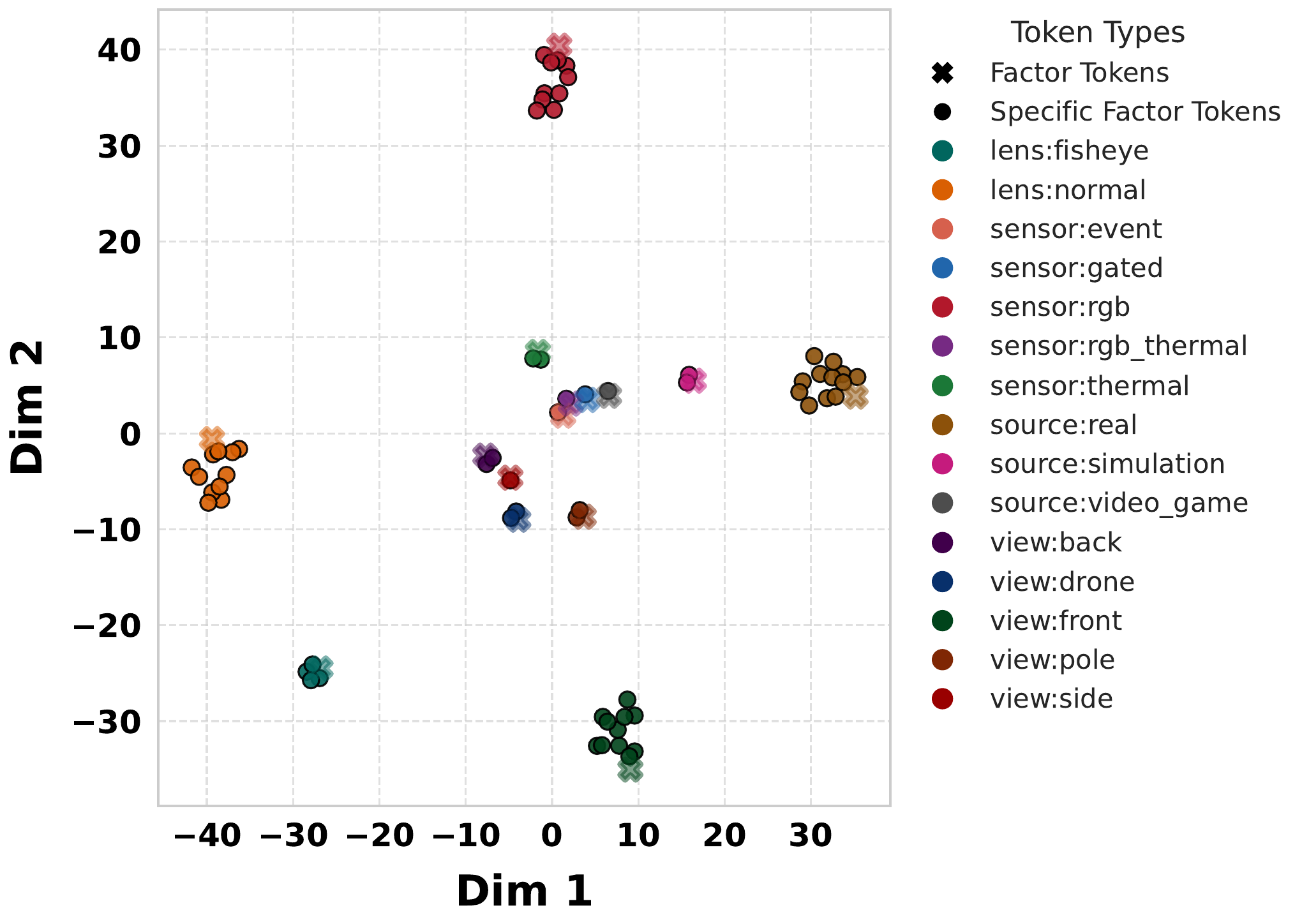}
    \end{minipage}\hfill
    \begin{minipage}{0.48\textwidth}
        \centering
        \includegraphics[height=4cm,keepaspectratio]{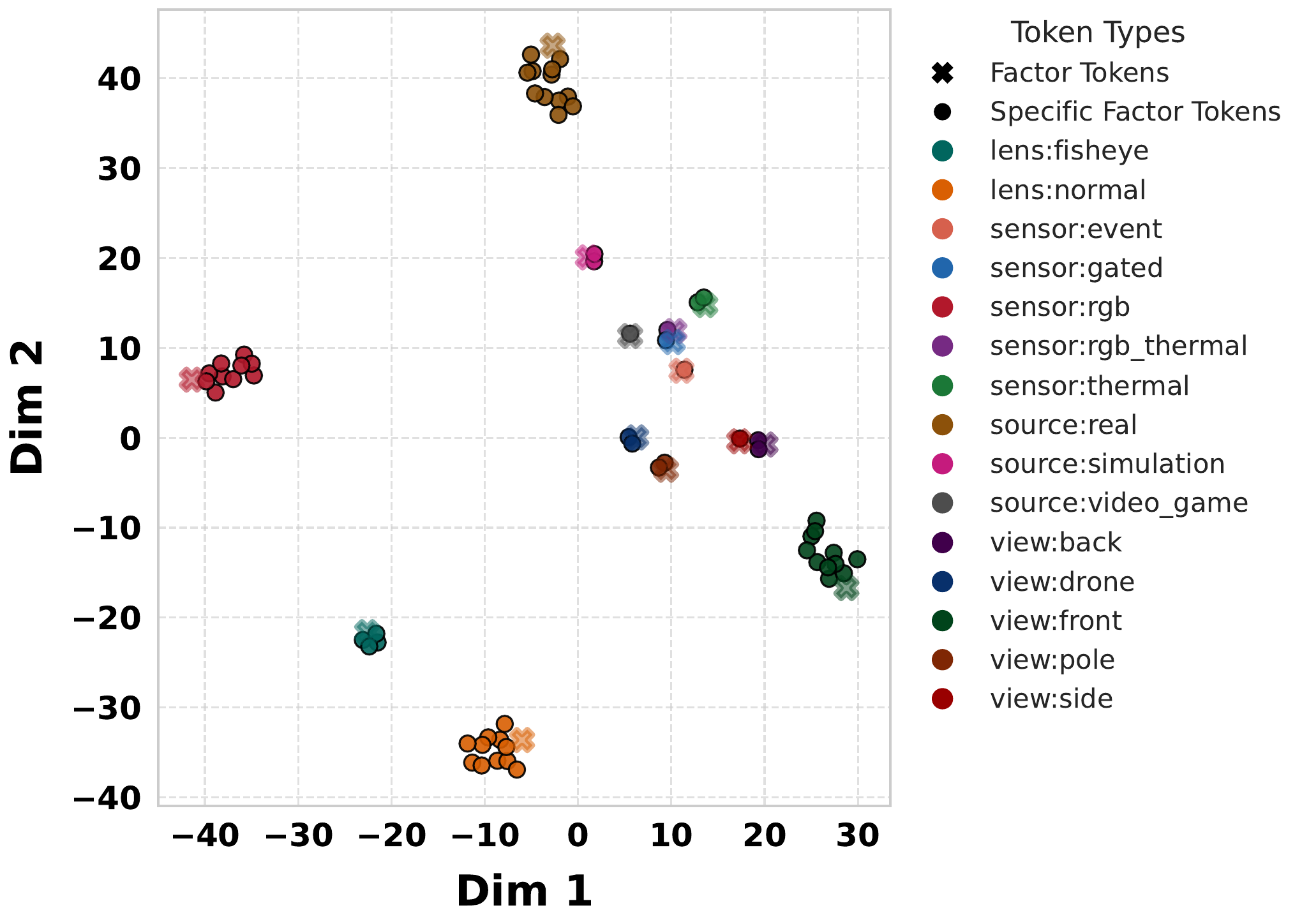}
    \end{minipage}
    \caption{\textbf{Embedding space visualization} across Text Encoder 1 (left) and Text Encoder 2 (right) using PCA~\cite{hotelling1933analysis,pearson1901liii} and t-SNE~\cite{maaten2008visualizing}.}
    \label{fig:embedding_plot_visualization}
\end{figure}

We visualize the learned factor embeddings for both of the text encoders of SDXL~\cite{podell2023sdxl} to analyze the relationship between general factors and dataset-specific factors. For each factor token, we extract its learned embedding vector for each of the corresponding text encoders. When multiple vectors are used to represent the same factor, we average the vectors to obtain a single representation. We apply dimensionality reduction to the learned embeddings of length 768, 1024~\cite{podell2023sdxl} to 10 using Principal Component Analysis (PCA)~\cite{hotelling1933analysis,pearson1901liii} to suppress noise and improve computational efficiency and then map them to a 2-dimensional space using t-distributed Stochastic Neighbor Embedding (t-SNE)~\cite{maaten2008visualizing} for visualization. Both general and specific factors are visualized jointly with different markers. The Fig.~\ref{fig:embedding_plot_visualization} shows that specific factors consistently cluster near their corresponding general factors across both of the text encoders. This indicates that specific factors act as a refinement of the general concept rather than as independent concepts.

\section{User Study}

\begin{figure}[h]
    \centering
    \begin{minipage}{0.8\linewidth}
        \begin{tcolorbox}[
            enhanced,      
            colback=white,    
            colframe=gray!30,   
            boxrule=0.5pt,  
            sharp corners, 
            drop shadow,  
            boxsep=0pt, left=0pt, right=0pt, top=0pt, bottom=0pt
        ]
            \includegraphics[width=\linewidth]{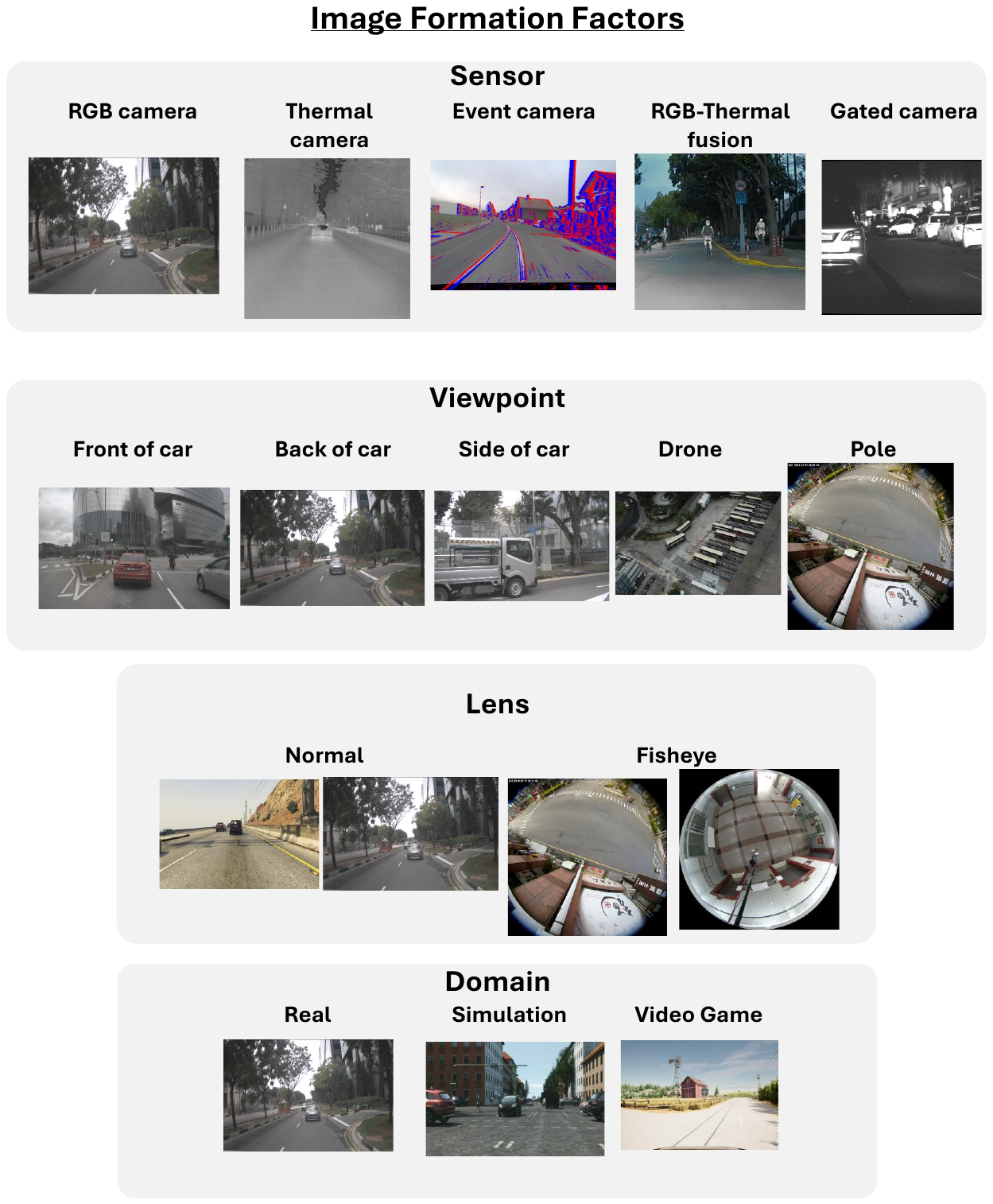}
        \end{tcolorbox}
    \end{minipage}
    \caption{\textbf{Reference Sheet} with example images for all four factors categories.}
    \label{fig:CheatSheet}
\end{figure}

The user study was designed to evaluate the quality and controllability of images generated by various methods. 
Each participant was asked $40$ questions, covering three categories: existing factor combinations, novel factor combinations and ControlNet guided controllability. The participants received a supplementary reference sheet as shown in Fig.~\ref{fig:CheatSheet}. They were also given a brief description to familiarize themselves with factors as follows:
\begin{itemize}
    \item \textbf{Sensor Type}
        \begin{itemize}
            \item \textbf{rgb}: Standard color camera images.
            \item \textbf{thermal}: Images where hotter objects appear brighter.
            \item \textbf{event}: Motion-triggered representations highlighting intensity changes (typically red/blue).
            \item \textbf{rgb-thermal}: Combined color and thermal information.
            \item \textbf{gated}: High-contrast images captured using gated imaging.
        \end{itemize}
        
    \item \textbf {Viewpoint}
        \begin{itemize}
            \item \textbf{front}: Camera facing forward.
            \item \textbf{back}: Camera facing backward.
            \item \textbf{side}: Camera mounted on the left or right side (typically on a vehicle).
            \item \textbf{drone}: Camera mounted on an aerial platform.
            \item \textbf{pole}: Camera mounted on a fixed pole.
        \end{itemize}
        
    \item \textbf{Lens Type}
        \begin{itemize}
            \item \textbf{normal}: Standard camera lens.
            \item \textbf{fisheye}: Wide field-of-view lens with edge distortion.
        \end{itemize}
        
    \item \textbf{Domain}
        \begin{itemize}
            \item \textbf{real}: Images captured in real-world environments.
            \item \textbf{simulation}: Rendered or simulated images.
            \item \textbf{video game}: Images captured from video game environments.
        \end{itemize}
\end{itemize}

\begin{figure}[t]
    \centering
    \includegraphics[width=1\linewidth]{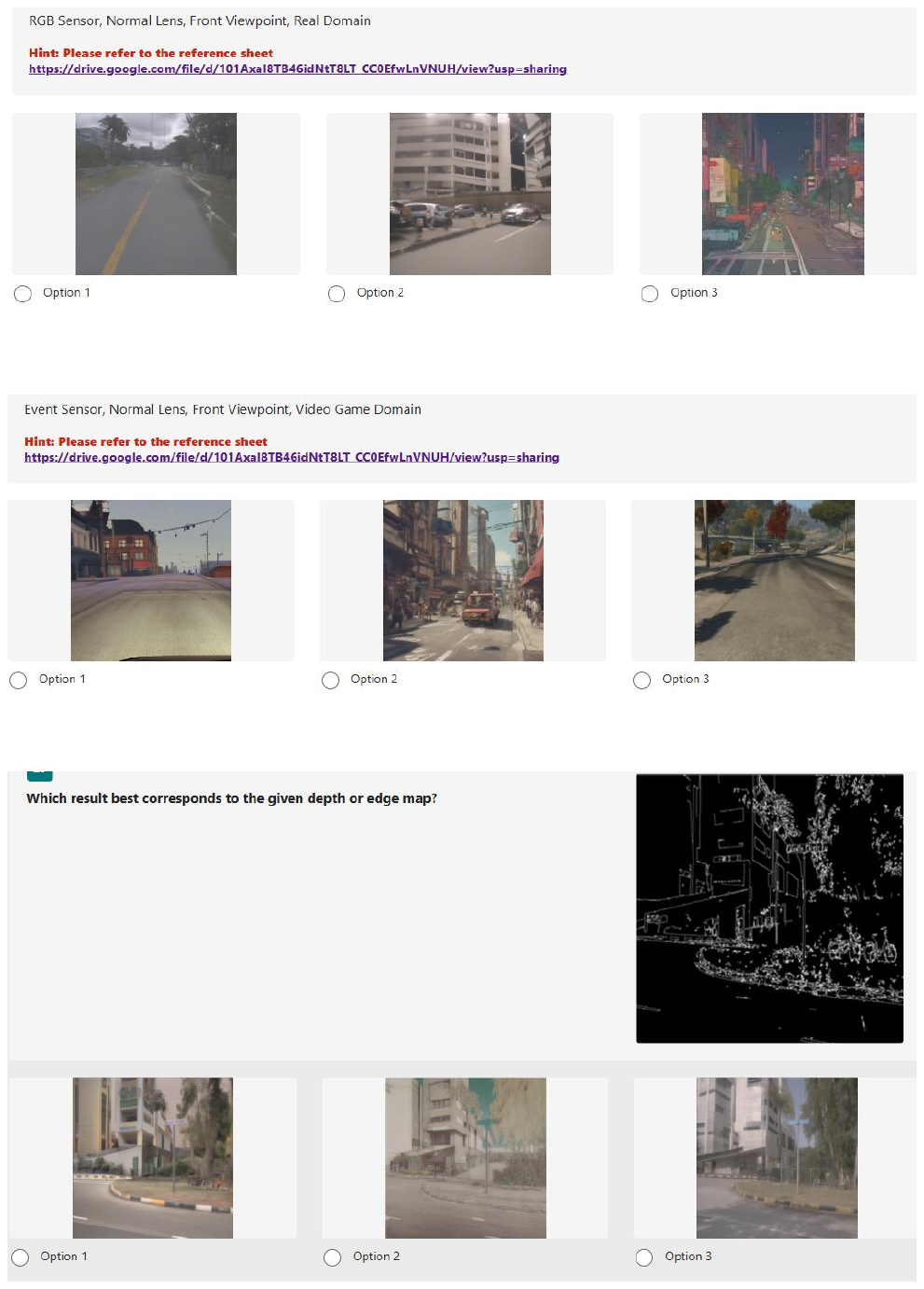}
    \caption{\textbf{User Study Questions.} Questions from categories: existing combination (top), novel combination (middle), controllability (bottom).}
    \label{fig:SampleQuestions}
\end{figure}

Each question states the factors used to generate the image, followed by one image generated by each of the compared methods. See Fig.~\ref{fig:SampleQuestions} for sample questions. Participants were instructed to select the image that best corresponds to the given combination of factors. To minimize bias, images were provided in random order and without method labels. A total of $27$ participants took part in the study. These participants were students and researchers from the computer vision field who were not involved in the project.

%% file: main.bbl
\begin{thebibliography}{10}
\providecommand{\url}[1]{\texttt{#1}}
\providecommand{\urlprefix}{URL }
\providecommand{\doi}[1]{https://doi.org/#1}

\bibitem{agnolucci2025isearle}
Agnolucci, L., Baldrati, A., Del~Bimbo, A., Bertini, M.: isearle: Improving textual inversion for zero-shot composed image retrieval. IEEE Transactions on Pattern Analysis and Machine Intelligence  (2025)

\bibitem{avrahami2023break}
Avrahami, O., Aberman, K., Fried, O., Cohen-Or, D., Lischinski, D.: Break-a-scene: Extracting multiple concepts from a single image. In: SIGGRAPH Asia. pp. 1--12 (2023)

\bibitem{baisa2025clip}
Baisa, N.L., Pallam, B., Jayavel, A.: Clip-handid: Vision-language model for hand-based person identification. arXiv preprint arXiv:2506.12447  (2025)

\bibitem{bijelic_seeing_2020}
Bijelic, M., Gruber, T., Mannan, F., Kraus, F., Ritter, W., Dietmayer, K., Heide, F.: Seeing {{Through Fog Without Seeing Fog}}: {{Deep Multimodal Sensor Fusion}} in {{Unseen Adverse Weather}}. In: {{CVPR}}. pp. 11679--11689 (Jun 2020). \doi{10.1109/CVPR42600.2020.01170}

\bibitem{bishop2006pattern}
Bishop, C.M., Nasrabadi, N.M.: Pattern recognition and machine learning, vol.~4. Springer (2006)

\bibitem{butt2024colorpeel}
Butt, M.A., Wang, K., Vazquez-Corral, J., van~de Weijer, J.: Colorpeel: Color prompt learning with diffusion models via color and shape disentanglement. In: ECCV. pp. 456--472. Springer (2024)

\bibitem{caesar_nuscenes_2020}
Caesar, H., Bankiti, V., Lang, A.H., Vora, S., Liong, V.E., Xu, Q., Krishnan, A., Pan, Y., Baldan, G., Beijbom, O.: {{nuScenes}}: {{A}} multimodal dataset for autonomous driving. arXiv preprint arXiv:1903.11027  (May 2020)

\bibitem{chen_amfd_2024}
Chen, Z., Qian, Y., Yang, X., Wang, C., Yang, M.: {{AMFD}}: {{Distillation}} via {{Adaptive Multimodal Fusion}} for {{Multispectral Pedestrian Detection}}. arXiv preprint arXiv:2405.12944  (May 2024). \doi{10.48550/arXiv.2405.12944}

\bibitem{dong2025dreamartist}
Dong, Z., Wei, P., Lin, L.: Dreamartist: Controllable one-shot text-to-image generation via positive-negative adapter: Z. dong et al. International Journal of Computer Vision  \textbf{133}(10),  7037--7053 (2025)

\bibitem{esser2024scaling}
Esser, P., Kulal, S., Blattmann, A., Entezari, R., M{\"u}ller, J., Saini, H., Levi, Y., Lorenz, D., Sauer, A., Boesel, F., et~al.: Scaling rectified flow transformers for high-resolution image synthesis. In: ICML. pp. 12606--12633. PMLR (2024)

\bibitem{teledyne_flir_free_2024}
FLIR, T.: {{FREE Teledyne FLIR Thermal Dataset}} for {{Algorithm Training}}. https://www.flir.com/oem/adas/adas-dataset-form/ (Accessed: 01122024)

\bibitem{gal2022image}
Gal, R., Alaluf, Y., Atzmon, Y., Patashnik, O., Bermano, A.H., Chechik, G., Cohen-or, D.: An image is worth one word: Personalizing text-to-image generation using textual inversion. In: ICLR (2022)

\bibitem{garibi2025tokenverse}
Garibi, D., Yadin, S., Paiss, R., Tov, O., Zada, S., Ephrat, A., Michaeli, T., Mosseri, I., Dekel, T.: Tokenverse: Versatile multi-concept personalization in token modulation space. ACM Transactions On Graphics (TOG)  \textbf{44}(4),  1--11 (2025)

\bibitem{gehrig_low-latency_2024}
Gehrig, D., Scaramuzza, D.: Low-latency automotive vision with event cameras. Nature  \textbf{629}(8014),  1034--1040 (May 2024). \doi{10.1038/s41586-024-07409-w}

\bibitem{gehrig_dsec_2021}
Gehrig, M., Aarents, W., Gehrig, D., Scaramuzza, D.: {{DSEC}}: {{A Stereo Event Camera Dataset}} for {{Driving Scenarios}}. IEEE Robot. Autom. Lett.  \textbf{6}(3),  4947--4954 (Jul 2021). \doi{10.1109/LRA.2021.3068942}

\bibitem{gochoo_fisheye8k_2023}
Gochoo, M., Otgonbold, M.E., Ganbold, E., Hsieh, J.W., Chang, M.C., Chen, P.Y., Dorj, B., Al~Jassmi, H., Batnasan, G., Alnajjar, F., Abduljabbar, M., Lin, F.P.: {{FishEye8K}}: {{A Benchmark}} and {{Dataset}} for {{Fisheye Camera Object Detection}}. In: {{CVPR Workshops}}. pp. 5305--5313 (Jun 2023). \doi{10.1109/CVPRW59228.2023.00559}

\bibitem{goodfellow2016deep}
Goodfellow, I., Bengio, Y., Courville, A., Bengio, Y.: Deep learning, vol.~1. MIT Press (2016)

\bibitem{ha2016hypernetworks}
Ha, D., Dai, A., Le, Q.V.: Hypernetworks. arXiv preprint arXiv:1609.09106  (2016)

\bibitem{hessel2021clipscore}
Hessel, J., Holtzman, A., Forbes, M., Le~Bras, R., Choi, Y.: Clipscore: A reference-free evaluation metric for image captioning. In: Proceedings of the 2021 conference on empirical methods in natural language processing. pp. 7514--7528 (2021)

\bibitem{heusel2017gans}
Heusel, M., Ramsauer, H., Unterthiner, T., Nessler, B., Hochreiter, S.: Gans trained by a two time-scale update rule converge to a local nash equilibrium. Advances in neural information processing systems  \textbf{30} (2017)

\bibitem{hotelling1933analysis}
Hotelling, H.: Analysis of a complex of statistical variables into principal components. Journal of educational psychology  \textbf{24}(6), ~417 (1933)

\bibitem{hu2022lora}
Hu, E.J., Shen, Y., Wallis, P., Allen-Zhu, Z., Li, Y., Wang, S., Wang, L., Chen, W., et~al.: Lora: Low-rank adaptation of large language models. ICLR  \textbf{1}(2), ~3 (2022)

\bibitem{johnson-roberson_driving_2017}
{Johnson-Roberson}, M., Barto, C., Mehta, R., Sridhar, S.N., Rosaen, K., Vasudevan, R.: Driving in the {{Matrix}}: {{Can Virtual Worlds Replace Human-Generated Annotations}} for {{Real World Tasks}}? arXiv preprint arXiv:1610.01983  (Feb 2017). \doi{10.48550/arXiv.1610.01983}

\bibitem{kansy2025reenact}
Kansy, M., Naruniec, J., Schroers, C., Gross, M., Weber, R.M.: Reenact anything: Semantic video motion transfer using motion-textual inversion. In: Proceedings of the Special Interest Group on Computer Graphics and Interactive Techniques Conference Conference Papers. pp. 1--12 (2025)

\bibitem{li2022blip}
Li, J., Li, D., Xiong, C., Hoi, S.: Blip: Bootstrapping language-image pre-training for unified vision-language understanding and generation. In: ICML. pp. 12888--12900. PMLR (2022)

\bibitem{Loshchilov2017DecoupledWD}
Loshchilov, I., Hutter, F.: Decoupled weight decay regularization. In: ICLR (2017)

\bibitem{maaten2008visualizing}
Maaten, L.v.d., Hinton, G.: Visualizing data using t-sne. Journal of machine learning research  \textbf{9}(Nov),  2579--2605 (2008)

\bibitem{motamed2024lego}
Motamed, S., Paudel, D.P., Van~Gool, L.: Lego: Learning to disentangle and invert personalized concepts beyond object appearance in text-to-image diffusion models. In: ECCV. pp. 116--133. Springer (2024)

\bibitem{neuwirth2025rico}
Neuwirth-Trapp, M., Bieshaar, M., Paudel, D.P., Van~Gool, L.: Rico: Two realistic benchmarks and an in-depth analysis for incremental learning in object detection. In: ICCVW. pp. 5153--5164 (2025)

\bibitem{nichol2022glide}
Nichol, A.Q., Dhariwal, P., Ramesh, A., Shyam, P., Mishkin, P., Mcgrew, B., Sutskever, I., Chen, M.: Glide: Towards photorealistic image generation and editing with text-guided diffusion models. In: ICML. pp. 16784--16804. PMLR (2022)

\bibitem{pearson1901liii}
Pearson, K.: Liii. on lines and planes of closest fit to systems of points in space. The London, Edinburgh, and Dublin philosophical magazine and journal of science  \textbf{2}(11),  559--572 (1901)

\bibitem{podell2023sdxl}
Podell, D., English, Z., Lacey, K., Blattmann, A., Dockhorn, T., M{\"u}ller, J., Penna, J., Rombach, R.: Sdxl: Improving latent diffusion models for high-resolution image synthesis. In: ICLR (2023)

\bibitem{rombach2022high}
Rombach, R., Blattmann, A., Lorenz, D., Esser, P., Ommer, B.: High-resolution image synthesis with latent diffusion models. In: CVPR. pp. 10684--10695 (2022)

\bibitem{ruiz2023dreambooth}
Ruiz, N., Li, Y., Jampani, V., Pritch, Y., Rubinstein, M., Aberman, K.: Dreambooth: Fine tuning text-to-image diffusion models for subject-driven generation. In: CVPR. pp. 22500--22510 (2023)

\bibitem{saharia2022photorealistic}
Saharia, C., Chan, W., Saxena, S., Li, L., Whang, J., Denton, E.L., Ghasemipour, K., Gontijo~Lopes, R., Karagol~Ayan, B., Salimans, T., et~al.: Photorealistic text-to-image diffusion models with deep language understanding. Advances in neural information processing systems  \textbf{35},  36479--36494 (2022)

\bibitem{salimans2016improved}
Salimans, T., Goodfellow, I., Zaremba, W., Cheung, V., Radford, A., Chen, X.: Improved techniques for training gans. Advances in neural information processing systems  \textbf{29} (2016)

\bibitem{schneider_timodataset_2022}
Schneider, P., Anisimov, Y., Islam, R., Mirbach, B., Rambach, J., Stricker, D., Grandidier, F.: {{TIMo}}---{{A Dataset}} for {{Indoor Building Monitoring}} with a {{Time-of-Flight Camera}}. Sensors  \textbf{22}(11), ~3992 (May 2022). \doi{10.3390/s22113992}

\bibitem{shentu2024attencraft}
Shentu, J., Watson, M., Al~Moubayed, N.: Attencraft: Attention-guided disentanglement of multiple concepts for text-to-image customization. CoRR  (2024)

\bibitem{simeoni2025dinov3}
Sim{\'e}oni, O., Vo, H.V., Seitzer, M., Baldassarre, F., Oquab, M., Jose, C., Khalidov, V., Szafraniec, M., Yi, S., Ramamonjisoa, M., et~al.: Dinov3. arXiv preprint arXiv:2508.10104  (2025)

\bibitem{sohn2023styledrop}
Sohn, K., Ruiz, N., Lee, K., Chin, D.C., Blok, I., Chang, H., Barber, J., Jiang, L., Entis, G., Li, Y., et~al.: Styledrop: text-to-image generation in any style. In: Proceedings of the 37th International Conference on Neural Information Processing Systems. pp. 66860--66889 (2023)

\bibitem{sun_shift_2022}
Sun, T., Segu, M., Postels, J., Wang, Y., Van~Gool, L., Schiele, B., Tombari, F., Yu, F.: {{SHIFT}}: {{A Synthetic Driving Dataset}} for {{Continuous Multi-Task Domain Adaptation}}. In: {{CVPR}}. pp. 21339--21350 (Jun 2022). \doi{10.1109/CVPR52688.2022.02068}

\bibitem{vinker2023concept}
Vinker, Y., Voynov, A., Cohen-Or, D., Shamir, A.: Concept decomposition for visual exploration and inspiration. ACM Transactions on Graphics (TOG)  \textbf{42}(6),  1--13 (2023)

\bibitem{wang2025multi}
Wang, K., Yang, F., Raducanu, B., van~de Weijer, J.: Multi-class textual-inversion secretly yields a semantic-agnostic classifier. In: WACV. pp. 4400--4409. IEEE (2025)

\bibitem{wei2023elite}
Wei, Y., Zhang, Y., Ji, Z., Bai, J., Zhang, L., Zuo, W.: Elite: Encoding visual concepts into textual embeddings for customized text-to-image generation. In: ICCV. pp. 15943--15953 (2023)

\bibitem{wei2025personalized}
Wei, Y., Zheng, Y., Zhang, Y., Liu, M., Ji, Z., Zhang, L., Zuo, W.: Personalized image generation with deep generative models: A decade survey. arXiv preprint arXiv:2502.13081  (2025)

\bibitem{de2024medical}
de~Wilde, B., Saha, A., de~Rooij, M., Huisman, H., Litjens, G.: Medical diffusion on a budget: Textual inversion for medical image generation. In: Medical Imaging with Deep Learning. pp. 1687--1706. PMLR (2024)

\bibitem{wrenninge_synscapes_2018}
Wrenninge, M., Unger, J.: Synscapes: {{A Photorealistic Synthetic Dataset}} for {{Street Scene Parsing}}. arXiv preprint arXiv:1810.08705  (Oct 2018). \doi{10.48550/arXiv.1810.08705}

\bibitem{xu2024dreamanime}
Xu, C., Xu, Y., Zhang, H., Xu, X., He, S.: Dreamanime: Learning style-identity textual disentanglement for anime and beyond. IEEE Transactions on Visualization and Computer Graphics  (2024)

\bibitem{xu2025cusconcept}
Xu, Z., Hao, S., Han, K.: Cusconcept: Customized visual concept decomposition with diffusion models. In: WACV. pp. 3678--3687. IEEE (2025)

\bibitem{yang_large-scale_2023}
Yang, L., Li, L., Xin, X., Sun, Y., Song, Q., Wang, W.: Large-{{Scale Person Detection}} and {{Localization}} using {{Overhead Fisheye Cameras}}. arXiv preprint arXiv:2307.08252  (Jul 2023). \doi{10.48550/arXiv.2307.08252}

\bibitem{yang2024pedestrian}
Yang, Z., Wu, D., Wu, C., Lin, Z., Gu, J., Wang, W.: A pedestrian is worth one prompt: Towards language guidance person re-identification. In: CVPR. pp. 17343--17353 (2024)

\bibitem{yogamani_woodscape_2021}
Yogamani, S., Hughes, C., Horgan, J., Sistu, G., Varley, P., O'Dea, D., Uricar, M., Milz, S., Simon, M., Amende, K., Witt, C., Rashed, H., Chennupati, S., Nayak, S., Mansoor, S., Perroton, X., Perez, P.: {{WoodScape}}: {{A}} multi-task, multi-camera fisheye dataset for autonomous driving. arXiv preprint arXiv:1905.01489  (Jul 2021). \doi{10.48550/arXiv.1905.01489}

\bibitem{yu_bdd100k_2020}
Yu, F., Chen, H., Wang, X., Xian, W., Chen, Y., Liu, F., Madhavan, V., Darrell, T.: {{BDD100K}}: {{A Diverse Driving Dataset}} for {{Heterogeneous Multitask Learning}}. arxiv preprint arXiv:1805.04687  (Apr 2020)

\bibitem{zhang2023adding}
Zhang, L., Rao, A., Agrawala, M.: Adding conditional control to text-to-image diffusion models. In: ICCV. pp. 3836--3847 (2023)

\bibitem{zhang2018unreasonable}
Zhang, R., Isola, P., Efros, A.A., Shechtman, E., Wang, O.: The unreasonable effectiveness of deep features as a perceptual metric. In: Proceedings of the IEEE conference on computer vision and pattern recognition. pp. 586--595 (2018)

\bibitem{zhang2024attention}
Zhang, Y., Yang, M., Zhou, Q., Wang, Z.: Attention calibration for disentangled text-to-image personalization. In: CVPR. pp. 4764--4774 (2024)

\bibitem{zhong2025mod}
Zhong, W., Yang, H., Liu, Z., He, H., He, Z., Niu, X., Zhang, D., Li, G.: Mod-adapter: Tuning-free and versatile multi-concept personalization via modulation adapter. arXiv preprint arXiv:2505.18612  (2025)

\bibitem{zhu_detection_2022}
Zhu, P., Wen, L., Du, D., Bian, X., Fan, H., Hu, Q., Ling, H.: Detection and {{Tracking Meet Drones Challenge}}. PAMI  \textbf{44}(11),  7380--7399 (Nov 2022). \doi{10.1109/TPAMI.2021.3119563}

\end{thebibliography}
